# An Input-Frugal Deep Learning Framework for Weather-Driven National Crop-Yield Forecasting: A Case Study of Brazilian Soybean

Fernando Dupin da Cunha Mello[a] (fdcmello@gmail.com), Prashant Kumar[b, c, d] (pkumar@surrey.ac.uk), and Erick G. Sperandio Nascimento[a, b, c, d,*] (erick.sperandio@surrey.ac.uk)

[a] Stricto Sensu Department, SENAI CIMATEC University, Salvador, Bahia, Brazil

[b] Global Centre for Clean Air Research (GCARE), School of Engineering, Civil and Environmental Engineering, Faculty of Engineering and Physical Sciences, University of Surrey, Guildford GU2 7XH, United Kingdom

[c] Institute for Sustainability, University of Surrey, Guildford, GU2 7XH, United Kingdom

[d] Surrey Institute for People-Centred AI, Faculty of Engineering and Physical Sciences, University of Surrey, Guildford GU2 7XH, UK

*Corresponding author email: erick.sperandio@surrey.ac.uk

## Abstract

Reliable, timely crop-yield forecasts are essential for market stability and risk management, yet many approaches rely on costly or hard-to-scale inputs. We present a frugal, transferable, and architecture-agnostic deep learning framework that uses routine weather as the only time-varying input plus two lightweight static context inputs (crop year and an agro-environmental label) to capture long-run change and regional heterogeneity, while supporting multiple sequence encoders under identical data requirements. Using a 20-season Brazilian soybean case study (2001/02–2020/21) with leave-one-year-out cross-validation, we benchmark MLP, CNN, LSTM, CNN–LSTM, a Transformer encoder and the Mamba state-space model against linear ridge regression and a five-year moving-average “farmer” baseline. All deep learning variants outperform ridge, and all sequential encoders surpass the non-sequential MLP. The Transformer achieves the best national accuracy (RMSE 149 kg ha$^{-1}$; rRMSE 5.3%; $R^2$ = 0.784), reducing error by 47.6% relative to the farmer baseline. In-season forecasts improve monotonically from early- to late-season issuance, reaching approximately 50% lower error than the baseline at the latest forecast point. Ablations indicate that the agro-environmental label and spatial instance expansion (multiple grid-node weather sequences per municipality–year) contribute positively without increasing input complexity. SHAP diagnostics suggest crop year explains most of the long-run trajectory, whereas within-season weather and agro-environmental context primarily drive interannual deviations, with moisture/cloud and thermal-demand variables dominating. Overall, the framework is straightforward to deploy across other crops and geographic regions and is naturally compatible with operational weather forecasts for routine monitoring.

# 1 Introduction

Global food security remains a persistent and complex challenge, aggravated by climate variability and commodity market volatility. Agricultural production is highly sensitive to weather fluctuations, which can disrupt supply chains and trigger price oscillations in staple crops (FAO *et al.*, 2024). These dynamics directly affect the four pillars of food security, availability, access, utilisation, and stability, and strengthen the case for forecasting tools that deliver reliable, timely information to stakeholders at multiple scales, consistent with United Nations Sustainable Development Goal target 2.C (United nations, 2016)

Traditional yield-forecasting approaches, whether statistical/econometric or process-based, face important limitations. Statistical models often depend on long, consistent time series that can be unavailable—or too coarse—at municipal scales, and they struggle to capture non-linear crop–weather responses (Mathieu and Aires, 2016; Qin *et al.*, 2024). Process-based models provide biophysical representations but require detailed local inputs on soil properties (Wimalasiri *et al.*, 2020), management practices (Timlin, Paff and Han, 2024), and cultivar traits (Jones *et al.*, 2003), which are often not readily available or straightforward to scale to national and global contexts. Operational systems can therefore face a trade-off between physical realism and the practical availability of consistent inputs at scale.

In recent years, machine learning and deep learning have shown strong potential to model complex, non-linear relationships in crop forecasting, with studies demonstrating competitive performance using remote sensing (You *et al.*, 2017; Ma *et al.*, 2021) and environmental predictors (Bloh, von *et al.*, 2023; Khaki and Wang, 2019) and reviews cataloguing a rapidly expanding evidence base (Oikonomidis, Catal and Kassahun, 2023). Systematic reviews further emphasise both the breadth of modelling approaches and the sensitivity of reported skill to data modality, spatial scale, and evaluation design (Klompenburg, van, Kassahun and Catal, 2020). Within soybean, recent work illustrates that strong accuracy can be achieved via remote-sensing-intensive pipelines and high-resolution yield mapping, often coupled with climate covariates and machine learning (Schwalbert *et al.*, 2020; Silva Fuzzo *et al.*, 2020; Song *et al.*, 2022a). These advances highlight the potential of modern machine-learning pipelines, but they also motivate closer attention to design choices that affect operational feasibility, reproducibility, and forecast readiness.

Despite this progress, three gaps remain particularly relevant for operationally useful yield forecasting at national and other aggregate scales.

1- High-performing pipelines often rely on complex or latency-prone inputs (for example, remote-sensing streams or detailed management and soil datasets), which can limit reproducibility and scalability across countries and crops (Schwalbert *et al.*, 2020; Silva Fuzzo *et al.*, 2020; Song *et al.*, 2022a).
2- Architecture comparisons are frequently confounded by differences in inputs, fusion strategies, and evaluation protocols, making it difficult to attribute gains to the sequence encoder itself (Klompenburg, van, Kassahun and Catal, 2020).
3- Many approaches are not explicitly forecast-ready, meaning that they are not designed to incorporate operational weather forecasts for in-season updating and early decision-making. Instead, they are used mainly for end-of-season estimation or mapping.

These gaps motivate a framework that prioritises reproducibility, enables fair encoder comparisons under identical inputs, and is explicitly designed for forecast-ready deployment. To operationalise these aims, this research proposes a scalable deep learning pipeline for crop-yield forecasting that relies on two classes of globally accessible and frugal inputs:

- High-resolution meteorological variables derived from reanalysis-based agrometeorological datasets.
- A simplified agro-environmental context labelling of municipalities. This functions as a proxy for difficult-to-observe agronomic heterogeneity—including soil type, management practices, and cultivar choices—enabling the model to capture persistent regional differences without explicitly collecting those datasets.

Together, these inputs allow the model to learn both temporal weather dynamics and persistent spatial patterns while minimising dependence on complex local information. In addition, we introduce a spatial instance expansion (SIE) strategy that exploits the fine spatial resolution of reanalysis/agrometeorological dataset AgERA5 (Boogaard *et al.*, 2020): multiple meteorological grid nodes located within each municipality are linked to the same observed municipal yield for the corresponding season. This procedure multiplies the number of training samples, increases dataset diversity, and reduces susceptibility to overfitting, while preserving a consistent yield reference at the municipality scale.

Complementary studies in Brazil have demonstrated the feasibility of climate-driven machine-learning yield prediction at sub-national scales, including state- and municipal-level analyses (Santos, dos *et al.*, 2022; Santos, dos, Santos, dos and Rolim, 2021; Torsoni *et al.*, 2023). Extending this logic, we propose an operational framework for aggregate-scale forecasting that relies only on routinely available weather inputs. We shift the emphasis to national-scale forecasting, evaluating models primarily on harvested-area-weighted national aggregates that align with operational monitoring and with the scale at which production shocks are most relevant to market impacts and commodity price formation (Summer and Mueller, 1989; Mello, Kumar and Sperandio Nascimento, 2024). We demonstrate the approach in a soybean case study for Brazil, a rigorous testbed given the crop's wide distribution across diverse biomes and agro-ecological zones (i.e., production environments stratified by climate and edaphic constraints) (FAO and IIAS, 2021) and its pronounced interannual weather-driven variability. Moreover, Brazil is currently the world's leading producer and exporter of soybeans (USDA, 2025), further motivating both the choice of soybean as the focal crop and Brazil as the study setting.

Within this common, frugal weather-driven framework, we benchmark multiple temporal encoders (MLP, CNN, LSTM, CNN–LSTM, Transformer, and Mamba) under identical inputs and fusion strategy, and compare them against linear and operational baselines.

Beyond its immediate application to soybean in Brazil, the broader contribution of this research is replicability. By restricting inputs to globally available meteorological data and a simple categorical labelling of production environments—aligned with established agro-ecological zoning concepts (FAO and IIAS, 2021)—this framework can be adapted to different crops and geographies with minimal bespoke data collection. In this way, the method constitutes a frugal yet effective approach to yield forecasting that can support agricultural planning, market stability, and food security worldwide.

Specifically, this study makes four contributions:

- First, it proposes a frugal and replication-oriented yield-forecasting framework that uses routine weather as the only time-varying input plus lightweight static context, thereby supporting deployment across crops and regions with minimal additional data engineering.
- Second, it establishes a controlled benchmarking pipeline that compares alternative temporal encoders under identical inputs, fusion strategy, and a strict leave-one-year-out protocol, while also reporting benchmark-normalised performance against a replicable "farmer" baseline to support comparability beyond this study.
- Third, it delivers a forecast-ready design in which yield predictions can be driven by operational weather-forecast inputs rather than only realised weather observations, and evaluates this capability through progressively earlier in-season predictions.

- Fourth, it provides a SHAP-based model diagnostics that clarify how crop year, within-season weather, and the agro-environmental context label contribute to interannual yield variability and model performance.

# 2 Methods and Materials

## 2.1 Framework design: problem formulation, pre-processing, and spatial instance expansion

### 2.1.1 Problem formulation

We develop a replication-oriented, weather-driven deep learning framework for national-scale crop-yield forecasting. The framework is designed around a frugal input structure, combining routinely available meteorological time series with a minimal set of static covariates, so that it can be adapted across crops and geographies with limited additional engineering.

We cast crop-yield forecasting as a supervised regression problem in which models are trained on subnational administrative units and evaluated primarily after aggregation, with harvested-area-weighted national yield as the main reporting scale. The learning objective is to predict observed unit-level yields from within-season meteorological information and minimal static context, and then assess how well these predictions reproduce aggregate outcomes relevant to operational monitoring and market impacts.

To keep the framework input-frugal while capturing persistent differences across production environments and long-run yield change, we include two static context covariates alongside the within-season meteorological sequence.

1. An agro-environmental context label **(**$z_m$**)** encodes coarse, structural, and slow-changing differences across locations (e.g., soil and climate regimes, production environments), allowing the model to learn regime-specific weather–yield relationships without relying on extensive site-specific covariates.
2. A crop-year (season) identifier **(**$\tau_t$**)** acts as a parsimonious proxy for time-varying technological progress and management improvements that can induce sustained yield trends over time (Pardey and Alston, [s.d.]; Umburanas *et al.*, 2022). Rather than detrending the series and re-adding the trend ex post, as commonly done in the literature (Li *et al.*, 2023; von Bloh *et al.*, 2023), we provide $\tau_t$ directly as an input so that models can learn long-run yield gains jointly with interannual weather–yield co-variation and potential interactions with agro-environmental context. This choice offers a generic, replication-friendly treatment of trend by reducing reliance on separate detrending/retrending steps.

Together, these covariates provide minimal static context that complements the weather sequence while preserving the framework's replication-oriented design.

### 2.1.2 Temporal alignment and aggregation

Temporal alignment within the framework is governed by a user-defined crop calendar that maps each crop season to a consistent within-season window (e.g., sowing-to-harvest) and defines the time-step granularity at which meteorological predictors are represented. The framework is agnostic to this choice: the within-season sequence can be represented at daily, weekly, monthly (or other) resolution, provided that the calendar, window definition, and sequence length are applied consistently across seasons and locations.

Where meteorological data are available at a finer resolution than the modelling time step, temporal aggregation is treated as a configurable pre-processing step rather than a fixed design decision. In many yield-forecasting settings, aggregation is used to reduce high-frequency noise and emphasise climatological signals relevant for crop development (Maas, 1982), but the appropriate granularity depends on the crop, region, and forecasting objective.

The framework also supports truncated-input variants for earlier issuance (in-season forecasting): the same temporal alignment rules apply, but only the first (k) time steps of the within-season sequence are retained, enabling like-for-like comparisons of lead-time performance under a fixed modelling pipeline.

### 2.1.3 Spatial instance expansion (SIE)

Deep learning models can exploit complex, non-linear relations but typically require large and diverse training sets to generalise well (Wang *et al.*, 2024). For each administrative unit $m \in M_t$, we exploit a scale mismatch that is common in operational settings: yield is observed at the administrative-unit level, whereas meteorological inputs are available at finer spatial resolution. Because yield histories and the completeness of official statistics may be limited in some settings, the framework includes an optional spatial instance expansion scheme that leverages the fine spatial granularity of gridded meteorological datasets without generating synthetic data.

For each administrative unit $m$ and season $t$, we identify the set of grid nodes $G(m)$ whose centroids fall within the unit boundary. SIE is implemented by expanding each unit–season pair $(m, t)$ into multiple node–season training instances indexed by $g \in G(m)$: each instance carries the same unit-level yield target $y_{m,t}$ and the same static context for the unit and season, while the within-season meteorological sequence varies across nodes according to local conditions. This expansion yields $| G(m) |$ instances per unit–season, increases the effective sample size, and injects realistic within-unit spatial variability through physically consistent gridded fields, without introducing synthetic yield labels.

Because $|G(m)|$ can vary substantially across units and, in aggregate, across context classes (which may differ in spatial extent and unit density), naive instance expansion could over-represent contexts with denser grid-node coverage. To mitigate this, during model fitting we apply a context-wise cap on expanded instances: within each LOYO training fold, we limit the number of node–season instances per context class to the count observed in the smallest class (random subsampling with a fixed seed), so that the expanded training set remains approximately balanced across contexts while retaining within-context meteorological diversity. This balancing is applied only to the training data in each fold and does not affect the held-out test year.

The spatial instance expansion procedure has four key properties.

1. Target-label invariance: because the target does not vary across nodes within a unit–season, gradients encourage the model to focus on features that are stable across nodes (signal) and to discount purely local, non-systematic fluctuations (noise).
2. No synthetic target labels are introduced: all targets correspond to observed yields, and expanded instances differ only in their meteorological inputs, which are taken directly from physically consistent gridded fields rather than artificially perturbed features or fabricated yield values.
3. Split integrity and leakage control: all node-level instances originating from the same unit–season remain within the same split (train or test) throughout the leave-one-year-out (LOYO) cross-validation protocol (Section 2.4), preventing information leakage between training and evaluation sets.
4. Controlled contextual representation: the context-wise cap prevents the expanded training set from being dominated by contexts with denser grid-node coverage, directly addressing potential bias due to unequal numbers of node-level samples tied to the same observed unit-level target. During evaluation, predictions are aggregated back to unit–year yield, so results remain anchored to the unit target scale rather than the expanded node count.

### 2.1.4 Input representation

To formalise the common input structure used in the framework, we represent meteorological predictors using a gridded dataset (for example, a reanalysis product providing variables on a regular latitude–longitude grid, such as Copernicus/ECMWF AgERA5, ERA5(-Land), or NASA POWER). For each unit $m$ and season $t$ this dataset defines a set of grid nodes $g \in G(m)$. For node $g$ and season $t$, we construct a weather matrix $X_{g,t} \in \mathbb{R}^{TxP}$, where $T$ is the number of within-season time steps and $P$ is the number of meteorological predictors per time step.

Each dataset instance is a node–season tuple $(X_{g,t}, z_m, \tau_t, y_{m,t})$, where $z_m$ is a categorical agro-environmental context label for unit $m$, $\tau_t$ is a crop-year (season) identifier, and $y_{m,t}$ is the observed unit-level yield (kg ha$^{-1}$). The model's task is to learn a mapping $f: (X_{g,t}, z_m, \tau_t) \mapsto y_{m,t}$, taking node-level weather and unit/season context as inputs and predicting the unit-level yield target. The framework remains agnostic to the temporal aggregation window: $T$ simply denotes the number of within-season time steps per instance, and sequences may be defined at daily, weekly, or monthly resolution provided a consistent crop calendar and sequence length are specified.

## 2.2 Deep learning encoders within the framework

### 2.2.1 Common fusion scheme and evaluated architectures

Within the proposed framework, all models share the same input structure and fusion scheme; they differ only in how they encode the within-season weather sequence. We benchmark five temporal encoders – a 1D CNN, an LSTM, a CNN–LSTM hybrid, a Transformer encoder, and the Mamba state-space model – against a non-sequential MLP baseline. A schematic overview of this common fusion scheme is shown in Figure 1.

The weather branch is the only time-varying input. Each temporal architecture receives the within-season weather matrix $X_{g,t} \in \mathbb{R}^{T \times P}$(Section 2.1.4), where $T$ denotes the number of time steps under the user-defined crop calendar and chosen temporal granularity (e.g., daily, weekly, monthly) and $P$ is the number of meteorological predictors. Each encoder maps this sequence to a fixed-length weather embedding via its own encoder-specific mechanism (for example, the final hidden state for recurrent models or a pooled representation for attention- and convolution-based encoders)

In parallel, the agro-environmental context label $z_m$ (one-hot) and crop-year identifier $\tau_t$ (numeric) are passed through small two-layer dense projections to obtain compact static embeddings. We then concatenate [weather embedding | context projection | year projection] and feed this vector to a three-layer MLP regression head, which learns non-linear interactions between within-season weather patterns, structural location context, and long-run trend.

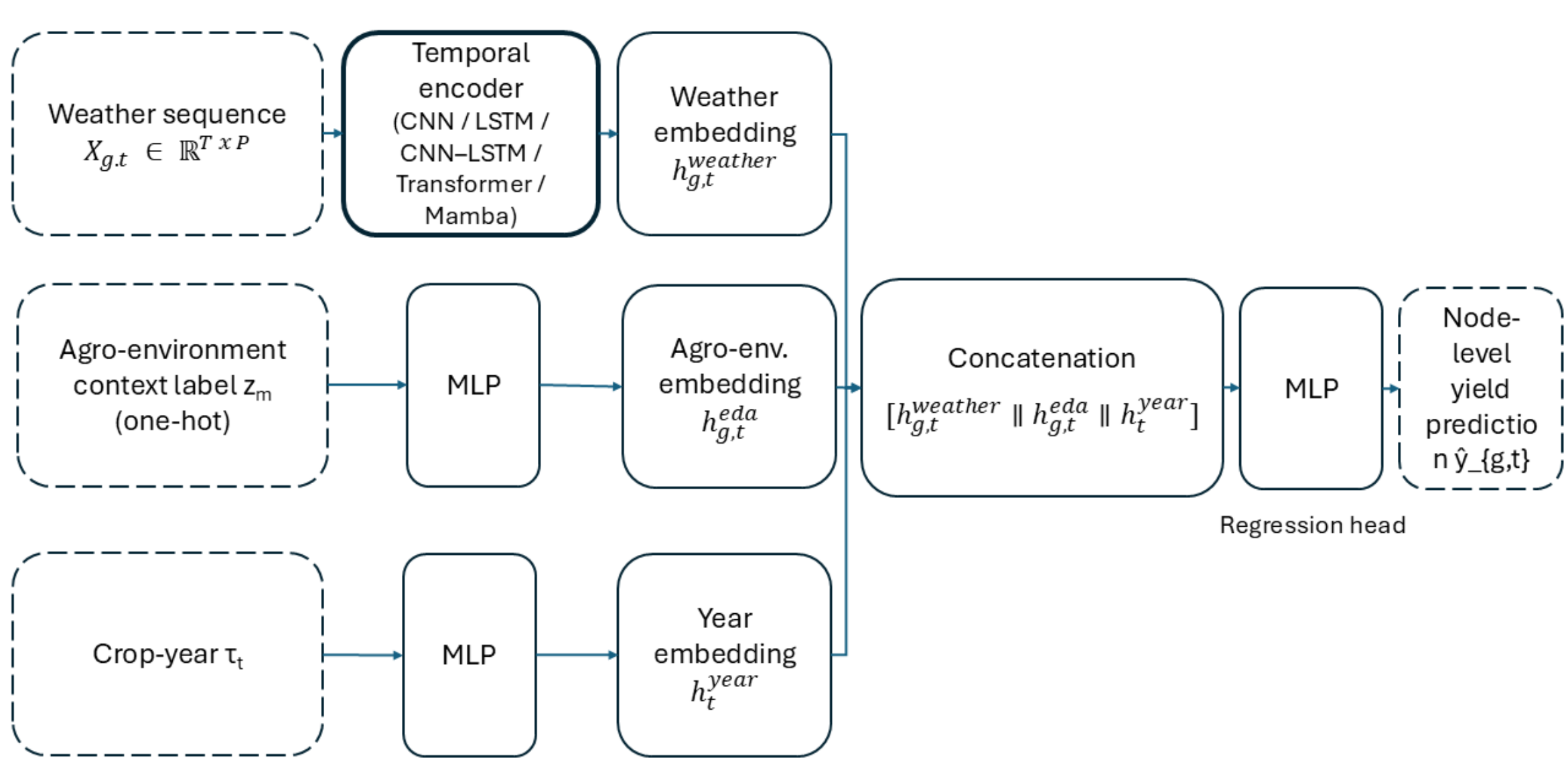

Figure 1: Schematic of the weather-driven deep learning framework and common fusion scheme. Dashed boxes denote observed quantities (inputs/outputs); whereas solid boxes denote trainable components and their latent embeddings. For each grid node $g$ within administrative unit $m$ and crop season $t$, the model ingests a within-season weather sequence (from a gridded meteorological dataset), an agro-environmental context label and a crop-year identifier. The temporal encoder can be instantiated as different architectures (CNN, LSTM, CNN–LSTM, Transformer or Mamba) while the static branches and regression head remain unchanged. This design allows a fair comparison of alternative sequence encoders under identical inputs and fusion strategy.

For the MLP baseline, the weather matrix $X_{g,t}$ is flattened (no temporal encoder) and concatenated directly with the same static projections before entering the regression head. This baseline ignores time-step order and temporal structure; any systematic improvement obtained by the sequence encoders under identical inputs can therefore be attributed to exploiting temporal dependencies rather than mere feature availability.

### 2.2.2 Transformer encoder (reference implementation)

Within the proposed framework, we designate a Transformer encoder as the primary reference model for detailed description, ablation experiments, in-season variants and explainability analyses. This choice is motivated by its ability to model long-range temporal dependencies through self-attention, its scalability to longer input sequences (e.g. weekly or daily weather) and its growing use as a standard architecture for sequence modelling. Performance comparisons between the Transformer and the other encoders are reported in Section 3.1. The structure of this Transformer-based weather encoder is illustrated schematically in Figure 2.

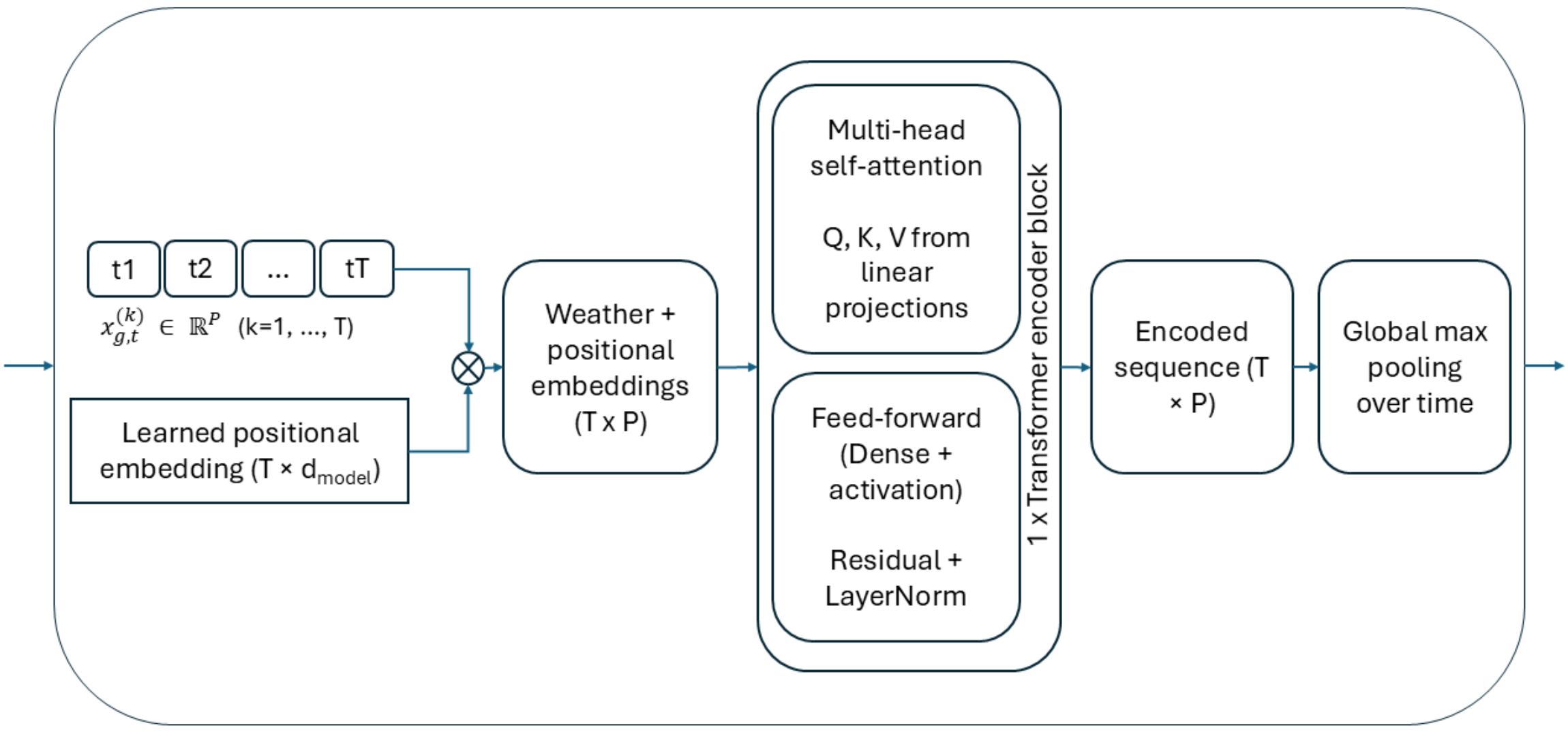

Figure 2: Transformer-based weather encoder within the proposed framework. Schematic of the weather branch instantiated with a Transformer encoder. Within-season gridded weather vectors $x_t \in \mathbb{R}^P$ (for $t = 1, \dots, T$, as defined by the user-specified crop calendar and temporal granularity) are treated as a short input sequence and combined with a learned positional embedding of size $T \times P$, yielding position-aware embeddings. This sequence is passed through a single Transformer encoder block comprising multi-head self-attention followed by a position-wise feed-forward subnetwork with residual connections and layer normalisation, producing an encoded sequence of shape $T \times d_{\text{model}}$ (with $d_{\text{model}} = P$ in this implementation). Global max pooling over time then generates a fixed-length weather embedding, which is subsequently fused with agro-environmental context and crop-year embeddings in the common fusion scheme (Figure 1).

For each grid node $g$ in administrative unit $m$ and crop season $t$, the weather input is the matrix $X_{g,t} \in \mathbb{R}^{T \times P}$, where each row corresponds to the $P$ meteorological predictors at one within-season time step. The model treats these rows as a short sequence of "tokens" representing the progression of the growing season. In our implementation, each time-step vector is represented in a space of dimension $d_{\text{model}}$, with $d_{\text{model}} = P$, and add a learned positional embedding of size $T \times d_{\text{model}}$ to the raw input matrix so that each time step has an associated position-dependent offset. The resulting embedded sequence, of shape $T \times d_{\text{model}}$, is then passed to the Transformer encoder. The time steps are always presented in the canonical order defined by the crop calendar (Section 2.1.2), so temporal position is conveyed jointly by the fixed ordering and the learned positional offsets.

The embedded sequence is then processed by a single Transformer encoder block, composed of a multi-head self-attention layer followed by a position-wise feed-forward network with residual connections and normalisation. In each block, self-attention computes, for every time step, a weighted combination of information from all time steps in the within-season sequence. Using the standard scaled dot-product attention mechanism, queries $Q$, keys $K$ and values $V$ are obtained from linear transformations of the embedded sequence, and attention is calculated as

$$\text{Attention}(Q, K, V) = \text{softmax}\left(\frac{QK^{\top}}{\sqrt{d_k}}\right)V, \tag{1}$$

where $d_k$ is the key dimension. Multi-head attention replicates this mechanism in parallel "heads", allowing the model to attend simultaneously to different aspects of the seasonal pattern (for instance, early-season heat versus late-season water stress). After the encoder block, we apply global max pooling along the temporal dimension to obtain a fixed-length weather embedding, which is subsequently fused with the context and crop-year embeddings in the common fusion scheme (Figure 1).

All key architectural and optimisation hyperparameters for the Transformer – including the number of attention heads, the key dimension, the width of the feed-forward sublayer, dropout rates, activation function, and learning-rate schedule – are selected during an initial development stage (Section 2.4) and then held fixed across all cross-validation folds. Full configuration details are provided in the Supplementary Material to facilitate replication and adaptation of the reference implementation to other crops and temporal resolutions within the same framework.

### 2.2.3 Other encoders within the framework

For completeness, we briefly summarise the other architectures instantiated within the same framework, beyond the Transformer reference model described in the previous section. All of them use the identical input representation, fusion scheme, and training protocol; only the temporal encoder branch changes.

- **MLP (non-sequential baseline).** A feed-forward network applied directly to the flattened weather matrix $X_{g,t}$, concatenated with $(z_m, \tau_t)$. Without any temporal inductive bias or explicit modelling of month-to-month order, it serves as a baseline for assessing the incremental value of sequence modelling.
- **1D CNN.** A one-dimensional convolutional network over the within-season time axis, with shared kernel filters capturing short intra-season patterns such as heat or dry spells. By exploiting local structure with relatively few parameters, the CNN typically improves parameter efficiency and reduces overfitting compared with the MLP on short sequences.
- **LSTM.** A stack of LSTM layers that encodes month-to-month dependencies via recurrent connections and parameter sharing across time. The sequence representation is taken from the final recurrent state, which summarises cumulative conditions across the season.

- **CNN–LSTM hybrid.** A hybrid encoder that first applies temporal convolutions to extract local patterns and then uses an LSTM to model longer-range dependencies, combining complementary inductive biases from convolutional and recurrent architectures.
- **Mamba (state-space model).** A selective, input-gated state-space encoder that updates its hidden state as a function of the current input, providing content-aware emphasis with linear-time complexity in sequence length (no quadratic attention). This makes it particularly attractive for potential extensions of the framework to higher-frequency weather inputs.

Together, these encoders and the Transformer reference model situate the proposed framework within a broad family of sequential models. The shared fusion scheme and input representation ensure that performance differences across architectures can be interpreted as arising from their respective temporal inductive biases, rather than from differences in data handling or covariate specification.

## 2.3 The case study

Brazilian soybean production provides a demanding and policy-relevant testbed for the proposed weather-driven yield-forecasting framework. Soybean is cultivated across all major Brazilian regions, spanning multiple biomes and a wide range of climatic regimes, which enables evaluation under diverse agro-meteorological conditions. The case study is designed to stress the framework's ability to learn yield–weather relationships across heterogeneous environments while reporting performance at the national aggregate level.

### 2.3.1 Study domain and unit of analysis

The framework is implemented at the municipality scale, using Brazilian soybean yields as the target variable for the 2001/02–2020/21 crop seasons. The study focuses on 10 major soybean-producing states (Bahia (BA), Goiás (GO), Maranhão (MA), Mato Grosso (MT), Mato Grosso do Sul (MS), Minas Gerais (MG), Paraná (PR), Rio Grande do Sul (RS), Santa Catarina (SC), and Tocantins (TO), covering 1,360 producing municipalities and accounting for more than 90% of national production in the reference season shown in Figure 3.

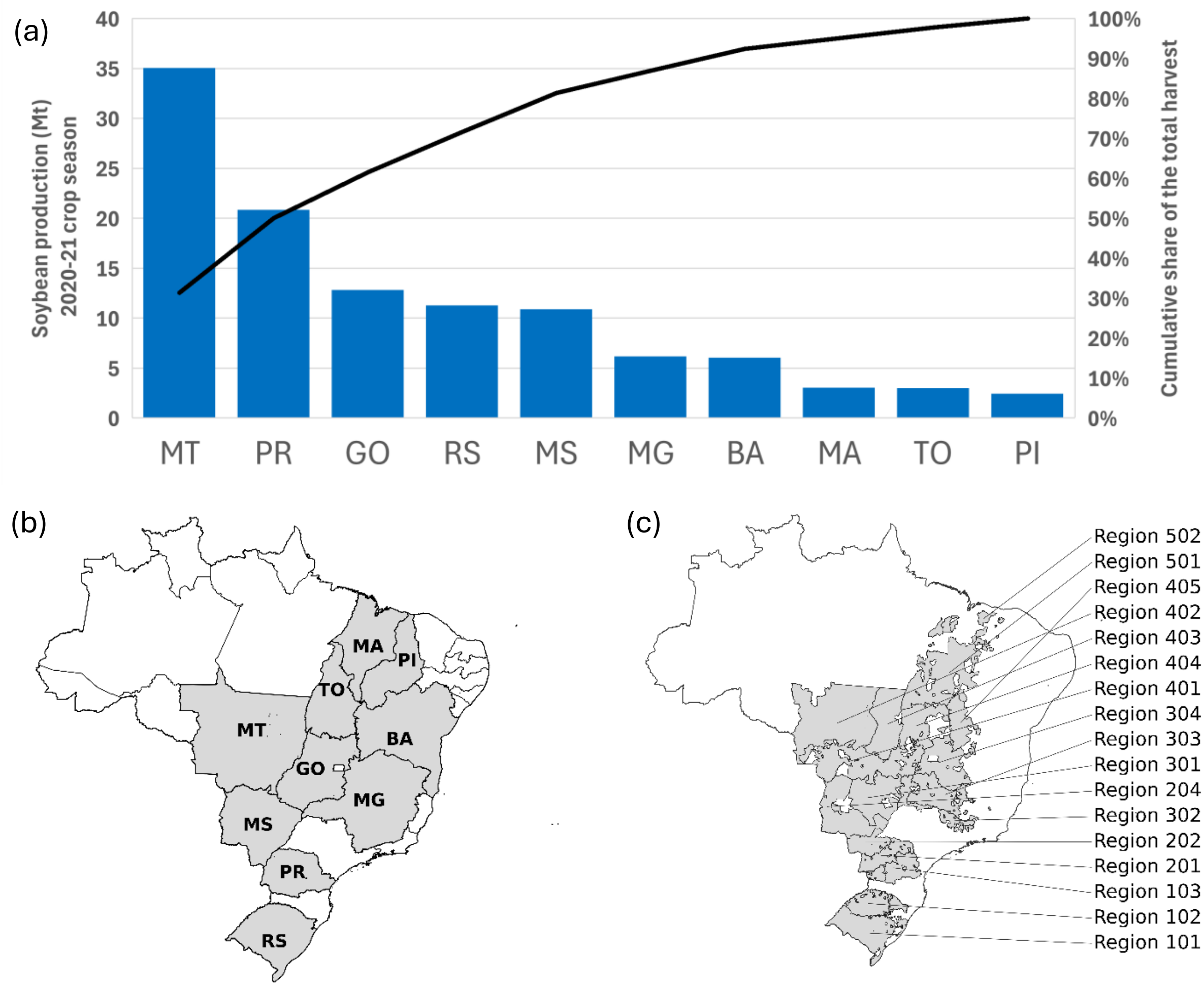

Figure 3: Study domain and production context. (a) Pareto chart of soybean production by state among the ten study states in 2020/21 (values in Mt, sorted descending). (b) Brazil with the ten participating states highlighted. (c) Clusters of soybean-producing municipalities grouped by edaphoclimatic regions (n = 17); outlines depict only the spatial envelope of producing municipalities within the study states; regions are used as the categorical context variable in the models.

### 2.3.2 Data sources

**Soybean yield and area (PAM/IBGE)**

Municipal soybean yields (kg ha$^{-1}$) are obtained from the *Produção Agrícola Municipal* (PAM) survey of the Brazilian Institute of Geography and Statistics (IBGE). For every municipality with recorded soybean production within the study states, we also extract planted and harvested areas, which are later used as weights when aggregating predictions to broader spatial units (state and national level).

**Meteorological predictors (AgERA5)**

Weather inputs are derived from AgERA5, a daily land-only reanalysis with approximately 0.1° × 0.1° spatial resolution. As shown in Table 1, from the broader indicator set, we retain ten variables with direct agronomic relevance (e.g., temperature, precipitation, humidity, radiation), excluding snow-related indicators as negligible for Brazilian soybean regions. Daily series are aligned to the soybean growing cycle and aggregated into within-season time steps; in this case study we use monthly aggregation and adopt September–April of each crop season as the reference window.

| Name | Units | Description |
|---|---|---|
| 10m wind speed | $m * s^{-1}$ | Mean wind speed at a height of 10 metres above the surface over the period 00h-24h local time. |
| 2m dewpoint temperature | K | Mean dewpoint temperature at a height of 2 metres above the surface over the period 00h-24h local time. The dew point is the temperature to which air must be cooled to become saturated with water vapor. In combination with the air temperature, it is used to assess relative humidity. |
| 2m relative humidity | % | Relative humidity at 06h, 09h, 12h. 15h, 18h (local time) at a height of 2 metres above the surface. This variable describes the amount of water vapour present in air expressed as a percentage of the amount needed for saturation at the same temperature. |
| 2m temperature | K | Air temperature at a height of 2 metres above the surface. |
| Cloud cover | Dimensionless | The number of hours with clouds over the period 00h-24h local time divided by 24 hours. |
| Liquid precipitation duration fraction | Dimensionless | The number of hours with precipitation over the period 00h-24h local time divided by 24 hours and per unit of area. Liquid precipitation is equivalent to the height of water that would accumulate if there were no infiltration, runoff, or evaporation. It is calculated using the ERA5 variable 'ptype' by determining the fraction of the day during which precipitation occurs in liquid form. |
| Precipitation flux | $mm * day^{-1}$ | Total volume of liquid water (mm3) precipitated over the period 00h-24h local time per unit of area (mm2), per day. |
| Solar radiation flux | $J * m^{-2} * day^{-1}$ | Total amount of energy provided by solar radiation at the surface over the period 00-24h local time per unit area and time. |
| Solid precipitation duration fraction | Dimensionless | The number of hours with solid precipitation (freezing rain, snow, wet snow, mixture of rain and snow, and ice pellets) over the period 00h-24h local time divided by 24 hours and per unit of area. It is calculated using the ERA5 variable 'ptype' by determining the fraction of the day during which precipitation occurs in solid form. |
| Vapour pressure | hPa | Contribution to the total atmospheric pressure provided by the water vapour over the period 00-24h local time per unit of time. |

Table 1: AgERA5 meteorological predictors used in the study.

### Agro-environmental context label (Brazilian edaphoclimatic regions)

To represent stable spatial heterogeneity without assembling detailed soil, management, or cultivar datasets, we define an agro-environmental context label as a compact categorical covariate. In this case study, the context label is instantiated using an official Brazilian edaphoclimatic regionalisation from Brazil's Ministry of Agriculture (MAPA). Each soybean-producing municipality is assigned to a single edaphoclimatic region (17 regions in total within the study domain; Figure 3c, matched to IBGE municipal boundaries. In the models, this context label conditions weather–yield relationships on relatively stable agro-environmental differences while keeping inputs frugal and facilitating replication.

### 2.3.3 Case-study input construction

**Temporal alignment and sequence definition**

Although the framework is agnostic to the temporal aggregation level in principle (Section 2.1.2), this case study instantiates the pipeline using monthly summaries aligned to the Brazilian soybean calendar. This choice is supported by evidence from soybean yield prediction studies indicating that moving from monthly to higher-frequency (e.g., daily) climate predictors does not necessarily improve predictive performance (Chen, Guilpart and Makowski, 2024). Using monthly summaries also provides an intentionally information-compressed representation of within-season weather; relative to weekly or daily inputs, it may obscure short-lived extremes and event timing, thereby offering a conservative test of whether the framework can extract predictive structure from frugal, widely available signals. For the core modelling experiments, we truncate the seasonal window to the first six months (September–February), reflecting that the national harvest is largely concluded by February and that later months are expected to add limited value for pre-harvest forecasting at the country level (von Bloh *et al.*, 2023). This yields $T = 6$ monthly steps per instance with $P = 10$ meteorological predictors per step (Table 1).

**Node-level representation and spatial instance expansion**

In this case study, spatial instance expansion (Section 2.1.3) is instantiated using AgERA5 meteorology at grid-node resolution and municipal yield observations. Each municipality–season record is linked to the set of AgERA5 grid nodes whose centroids fall within the municipal boundary, yielding node-level weather sequences $X_{g,t}$ paired with the same municipal yield target and context covariates. All balancing and split-integrity rules follow the framework specification in Section 2.1.3.

## 2.4 Training, validation, and hyperparameter tuning

All deep-learning models were trained using the same data pipeline, loss function, and validation strategy to ensure a fair comparison across encoders. For the main evaluation we adopted a leave-one-year-out (LOYO) cross-validation scheme over the 20 crop seasons from 2001/02 to 2020/21. In each fold, one crop season (t) was held out as a test year, and all remaining seasons were used for model fitting. Feature normalisation and any other input transformations were fitted using only data from the non-held-out years (training plus internal validation) within each fold and then applied unchanged to the corresponding test year, ensuring that no information from the held-out season leaked into model estimation or evaluation.

Within each LOYO fold, model weights were optimised on the training years and monitored using an internal validation split. Specifically, we created an 80/20 group-aware split of the training data at the municipality–year level (fixed random seed, identical partition across models) and used the resulting validation subset exclusively for early stopping and model selection within the fold. Training proceeded for up to a pre-defined maximum number of epochs, with early stopping triggered when the validation loss failed to improve for a fixed patience window; the best epoch according to validation performance was retained. The held-out year was never used for training or for early-stopping decisions.

Hyperparameters controlling model capacity and optimisation were selected in an initial development stage, prior to the LOYO evaluation. For each architecture (MLP, 1D CNN, LSTM, CNN–LSTM, Transformer and Mamba), we performed a structured search with Keras Tuner's BayesianOptimization algorithm (Snoek, Larochelle and Adams, 2012) on a separate development split constructed from the full 20-season dataset. This split provided a single training/validation partition used solely for hyperparameter selection and was not reused as an evaluation mechanism in the subsequent LOYO protocol. Each model family had its own architecture-specific search space; Table 2 summarises the hyperparameters and tested levels for the Transformer encoder as an illustrative example, and analogous sets were defined for the remaining deep-learning architectures.

For each architecture, the hyperparameter ranges were defined *a priori* as a pragmatic, literature-informed set centred on widely used defaults and constrained by computational budget. The goal was not to exhaustively optimise every model family, but to cover plausible capacity and optimisation regimes while keeping the comparison fair across encoders. Accordingly, search spaces included (i) a small number of discrete capacity levels (e.g., attention heads, key/query dimension, feed-forward width) spanning a moderate range to avoid extreme underfitting or over-parameterisation for a 20-season dataset; (ii) regularisation values (dropout) in a standard band used for sequence models; and (iii) learning rates covering an order of magnitude around common Adam defaults. This design ensures that Bayesian optimisation explores meaningful trade-offs (capacity vs regularisation vs optimisation) without drifting into configurations that are either implausible for the data regime or prohibitively expensive to train. Hyperparameters were fixed after the development-stage selection and were not adapted within LOYO, preserving the integrity of out-of-sample evaluation. For the Transformer, we restricted the encoder to a single block and modest head/dimension settings (Table 2) to match the short, monthly sequence length and to reduce overfitting risk, while still allowing attention to model cross-month interactions.

Candidate hyperparameter configurations for each architecture were first ranked by validation loss and then inspected using the full training history of loss curves, favouring settings that combined low validation error with a small and stable gap between training and validation loss. This joint criterion helped rule out clearly underfitting models (with consistently high losses) and overfitting models (with large or unstable train–validation gaps), even when their best validation epoch appeared competitive in isolation. Once a single configuration had been selected for each architecture, all hyperparameters were frozen and the models were retrained within the LOYO protocol. The Transformer and the other deep-learning models were trained with an Adam optimiser and mean squared error loss, reflecting the continuous nature of the yield target.

| **Hyperparameter** | **Description** | **Search space (tested levels)** |
|---|---|---|
| num_heads | Number of attention heads in multi-head self-attention | {2, **4**} |
| key_dim | Dimensionality of queries and keys per head | {**8**, 16, 32} |
| ff_dim | Width of the feed-forward sublayer in the encoder block | {64, 128, **256**} |
| dropout_rate | Dropout applied after attention and feed-forward sublayers | {**0.1**, 0.2, 0.3} (continuous in Keras Tuner with step 0.1) |
| activation | Activation function in the feed-forward sublayer | {**ReLU**, GELU} |
| learning_rate | Initial learning rate for the Adam optimiser | {$\mathbf{1\times10^{-3}}$, $5\times10^{-4}$, $1\times10^{-4}$} |

Table 2: Hyperparameter search space for the Transformer encoder. Search space used in the development-stage hyperparameter tuning of the Transformer-based weather encoder, implemented with Keras Tuner's BayesianOptimization. Each trial was trained on the same development split and evaluated by validation loss. The final configuration was selected by combining validation performance with inspection of training and validation loss curves. Search ranges were specified a priori based on standard practice and computational constraints. Boldface indicates the hyperparameter level selected for the final configuration used in the LOYO experiments.

This two-stage procedure (development-stage hyperparameter tuning followed by LOYO evaluation with fixed configurations) ensures that reported test results reflect genuine out-of-sample performance and that no hyperparameter "peeks" at the LOYO test years occur.

## 2.5 Prediction aggregation and baselines

Model outputs are generated at the node–season level. To evaluate performance at the spatial scales that matter operationally, we first "de-expand" predictions back to municipalities and then aggregate to state and national levels. All aggregation steps are deterministic and identical across architectures.

**Node-to-municipality aggregation (inverse of SIE):** Let $G(m)$ denote the set of AgERA5 grid nodes whose centroids fall within municipality $m$. For a given crop season $t$, the model produces node-level yield predictions $\{\hat{y}_{g,t}: g \in G(m)\}$. These are collapsed to a single municipal yield by simple averaging:

$$\hat{y}_{m,t} = \frac{1}{|G(m)|} \sum_{g \in G(m)} \hat{y}_{g,t}, \tag{2}$$

so that each municipality–season is represented by one predicted yield, consistent with the scale of the observed target $y_{m,t}$.

**Municipality-to-country aggregation:** Let $\mathcal{M}_t$ be the set of soybean-producing municipalities observed in season $t$, and $A_{m,t}$ the harvested soybean area in municipality $m$ and season $t$. Country-level (i.e. 10-state aggregate) yield is obtained as a harvested-area-weighted mean of municipal yields,

$$\hat{Y}_{c,t} = \frac{\sum_{m \in \mathcal{M}_t} \hat{y}_{m,t} A_{m,t}}{\sum_{m \in \mathcal{M}_t} A_{m,t}}, \tag{3}$$

where $c$ indexes the study "country". Predicted national production is computed as

$$\hat{Q}_{c,t} = \sum_{m \in \mathcal{M}_t} \hat{y}_{m,t} A_{m,t}. \tag{4}$$

State-level aggregates are computed analogously and used as intermediate diagnostics of spatial performance, with the Results section summarising their behaviour via the median and range of error metrics across states. Headline results are reported primarily for harvested-area-weighted national yields, while municipality-level errors are used mainly as internal diagnostics.

**Baselines:** We benchmark the deep-learning encoders against two non-ML baselines representing, respectively, a simple operational heuristic and a linear statistical model.

1. **Farmer baseline (five-year rolling mean):**

   For each held-out year $t$ in the LOYO evaluation, the "farmer" baseline predicts national yield as the mean of the previous five *observed* national yields. This low-complexity benchmark, widely used in crop-yield studies and operational outlooks (Iizumi *et al.*, 2021; Morales and Villalobos, 2023; USDA, 2023), depends only on historical yield information and can, in principle, be computed before sowing. It therefore serves as our main operational comparator for both end-of-season and in-season forecasts.

2. **Ridge-regression baseline (linear):**

   The ridge model is fitted at the municipality–season level using monthly weather summaries (AgERA5 predictors aggregated over nodes within each municipality), the edaphoclimatic one-hot label and the crop-year identifier as inputs. Features are standardised using only the non-held-out years, and a ridge penalty is applied to address multicollinearity among predictors. The regularisation strength is selected on the same development split used for deep learning hyperparameter tuning and then held fixed for all LOYO folds. Municipal ridge predictions are aggregated to state and national scales using the same harvested-area-weighted scheme as for the deep-learning models, ensuring comparability across all approaches.

## 2.6 Evaluation metrics and statistical analysis, and contextual benchmarking

Model performance is evaluated exclusively on held-out data from the LOYO protocol. For each fold, node-level predictions $\hat{y}_{g,t}$ are first aggregated to municipalities and then to state and national levels as described in Section 2.5, and error metrics are computed on these aggregates and pooled across all test years.

At the municipality scale, we define the prediction error for municipality $m$ and season $t$ as

$$e_{m,t} = \hat{y}_{m,t} - y_{m,t}, \tag{5}$$

where $\hat{y}_{m,t}$ is the predicted yield obtained by aggregating node-level outputs back to the administrative-unit scale under the spatial instance expansion (SIE) scheme and $y_{m,t}$ is the observed municipal yield. Over the union of all LOYO test folds, these errors can be summarised via standard metrics such as root mean squared error (RMSE), mean absolute error (MAE) and mean error (bias). In this study, municipal metrics are used primarily as internal diagnostics; the main quantitative comparisons are reported at state and national scales.

For national-scale yield, we work with the harvested-area-weighted aggregates $Y_{c,t}$ and $\hat{Y}_{c,t}$ (Section 3.5) and define national errors $E_{c,t} = \hat{Y}_{c,t} - Y_{c,t}$. From these we compute RMSE, MAE, bias and the relative RMSE (rRMSE):

$$\text{RMSE} = \sqrt{\frac{1}{T}\sum_{t} E_{c,t}^{2}}, \tag{6}$$

$$\text{MAE} = \frac{1}{T}\sum_{t} | E_{c,t} |, \tag{7}$$

$$\text{Bias} = \frac{1}{T}\sum_{t} E_{c,t}, \tag{8}$$

$$\text{rRMSE} = 100 \times \frac{\text{RMSE}}{\bar{Y}_c}, \tag{9}$$

where $T$ is the number of LOYO test years and $\bar{Y}_c$ is the mean observed national yield across those years. We also report the coefficient of determination

$$R^2 = 1 - \frac{\sum_t \left(Y_{c,t} - \hat{Y}_{c,t}\right)^2}{\sum_t \left(Y_{c,t} - \bar{Y}_c\right)^2}, \tag{10}$$

which quantifies the fraction of interannual variance in national yield explained by the predictions.

Comparative performance against the operational "farmer" baseline is quantified via a forecast skill score defined in terms of national RMSE. For a given model and forecast setting (end-of-season or in-season), we compute

$$\text{Skill} = 100 \times \left(1 - \frac{\text{RMSE}_{\text{model}}}{\text{RMSE}_{\text{farmer}}}\right), \tag{11}$$

where $\text{RMSE}_{\text{farmer}}$ is the RMSE of the five-year rolling-mean baseline evaluated on the same LOYO test years. Positive Skill values indicate improvements over the baseline (with, for example, Skill $= 50\%$ corresponding to a halving of RMSE), whereas negative values indicate performance worse than the baseline. For head-to-head ranking of encoders at the national scale (e.g. Transformer vs LSTM), we further compare year-paired absolute errors using a two-sided Wilcoxon signed-rank test based on the $T = 20$ LOYO test years.

To examine spatial consistency, we also evaluate models at the state scale. Municipal predictions are aggregated to state–year using harvested-area weights, and state-level RMSE, MAE and bias are computed by pooling errors across LOYO years for each state (n = 20 per state). In the Results, we report the per-state diagnostics for the reference Transformer to assess whether national performance could be explained by regional error cancellation.

The Supplement additionally reports country-level explained variance (EV) and a normalised RMSE (NRMSE), defined respectively as

$$\text{EV} = 1 - \frac{\text{Var}\left(Y_{c,t} - \hat{Y}_{c,t}\right)}{\text{Var}\left(Y_{c,t}\right)}, \tag{12}$$

$$\text{NRMSE} = 100 \times \frac{\text{RMSE}}{\text{SD}(Y_{c,t})}, \quad (13)$$

to provide complementary views of national-scale performance.

Uncertainty in RMSE and rRMSE is quantified via non-parametric bootstrap rather than closed-form approximations, which would require strong distributional assumptions on errors and large-sample normality. We treat the pool of out-of-sample prediction–observation pairs across all LOYO folds as the target of inference and apply a bootstrap at the year level: each resample consists of $T$ years drawn with replacement from the 20-year test set, and the corresponding metrics are recomputed on the resampled series. We report 95% confidence intervals using the bias-corrected and accelerated (BCa) method (Efron, 1987) with 10 000 resamples (Davison and Hinkley, 1997), which is robust to moderate bias and non-normality. The resulting intervals reflect uncertainty due to interannual and spatial variability, as well as the variation induced by using different training windows in each LOYO fold, but they remain conditional on the chosen modelling procedure: they do not attempt to incorporate additional uncertainty from re-fitting networks under alternative hyperparameter choices or random initialisations, which would require a fully unconditional bootstrap that is computationally prohibitive for the deep-learning architectures considered here. This conditional-on-procedure strategy treats the fitted model or cross-validation procedure as fixed when obtaining confidence intervals for predictive performance by bootstrap resampling of out-of-sample predictions (e.g. Baseer *et al.*, 2025; Jahan, Cerral and Astitha, 2024; Lee *et al.*, 2025; Nugraha *et al.*, 2025), and is consistent with methodological guidance on bootstrap confidence intervals for predictive accuracy (Carpenter and Bithell, 2000).

As a sensitivity analysis to potential year-to-year dependence, we also assessed a moving-block bootstrap with short temporal blocks; conclusions were consistent with the year-wise BCa intervals and are reported in Supplementary Table 2.

**Contextual Benchmarking:** To contextualise the end-of-season national accuracy of our frugal Transformer model, we compiled a targeted set of Brazil-focused soybean-yield modelling studies identified via a Google Scholar search prioritising relevance to (i) Brazil soybean yield prediction, (ii) national or Brazil-wide scope, and (iii) reporting RMSE and/or a relative error metric. Studies were then screened for availability of full text and extractable performance metrics and validation descriptions. This benchmarking is intended to provide performance context rather than a systematic literature review.

## 2.7 Ablation study design

The ablation study quantifies the contribution of two independent design choices within the proposed framework: (i) the inclusion of an edaphoclimatic region label as a static covariate, and (ii) the use of Spatial Instance Expansion (SIE) scheme at training time. All ablations are conducted using the Transformer encoder as the reference architecture; for each ablation variant, data, splits, and hyperparameters are held fixed relative to the full-reference Transformer model.

**Edaphoclimatic region label ablation:** Each municipality is assigned to an edaphoclimatic region (soil–climate class) and encoded as a one-hot vector $z_m$, which is passed through a small dense subnetwork to form a static embedding (Section 2.2.1). To isolate the contribution of this label, we construct an ablated variant in which the edaphoclimatic input is zeroed while keeping the architecture, optimiser settings, and all other inputs (weather sequence and crop-year identifier) unchanged. The resulting model therefore differs from the reference Transformer only by the absence of explicit edaphoclimatic information.

**Spatial instance expansion ablation:** To assess the contribution of the spatial instance expansion step (Section 2.1.3), the reference framework is trained with SIE enabled, whereas in the ablation we disable SIE and train the Transformer model using a single deterministic grid node per municipality–season $(m, t)$. All other aspects of the training procedure (including early stopping and the evaluation protocol) are held constant.

**Evaluation set-up and reduced-data experiments:** All ablation models follow the same LOYO protocol, prediction aggregation, and evaluation metrics described in Sections 2.4–2.6. End-of-season ablation experiments are first run using the full set of available training years for each LOYO fold. In addition, reduced-data experiments are conducted for the SIE ablation study only: we retrain models on fixed fractions of the available training years (25 %, 50 %, 75 %, and 100 %), holding the LOYO test year, preprocessing pipeline, and random seeds constant. This design allows us to assess whether spatial instance expansion is particularly beneficial when historical data are scarce, while ensuring that any observed differences are attributable to the presence or absence of SIE rather than to changes in data splits or optimisation settings.

**Statistical comparisons:** For the edaphoclimatic input ablation, we compare paired municipality–year absolute errors between the full-reference Transformer and the no-edaphoclimatic variant. Let $|\, e_{m,t}^{\text{full}} \,|$ and $|\, e_{m,t}^{\text{no-eda}} \,|$ denote the absolute errors at municipality $m$ and year $t$ for the two models. We form differences $|\, e_{m,t}^{\text{no-eda}} \,| - |\, e_{m,t}^{\text{full}} \,|$ and estimate their mean using weighted least squares with harvested area as weights and cluster-robust standard errors with year as the clustering unit, to account for within-year dependence. A two-sided test of zero mean difference is conducted at $\alpha = 0.05$ (two-sided), and we reject the null of no effect when $p < 0.05$, assessing whether the edaphoclimatic label yields a statistically significant change in municipality-level absolute error.

For the SIE ablation, the focus is on national-scale accuracy and its behaviour under reduced-data settings. We compute annual absolute errors in national yield for the instance-expanded and non-expanded models and compare them using the paired Wilcoxon signed-rank test (Wilcoxon, 1945) across LOYO years. Because the question is directional ("does SIE reduce error when data are limited?"), we adopt a one-sided alternative hypothesis favouring lower absolute error for the SIE-enabled model. Results are summarised by training-fraction setting, providing a structured assessment of whether SIE offers a robustness benefit without degrading performance when all years are available.

## 2.8 In-season forecasting setup

In-season forecasting is implemented as a sequence of truncated-input variants of the reference Transformer model, designed to emulate progressively shorter lead times as the growing season unfolds. After completing the end-of-season evaluation and model comparison, we select the best-performing architecture (the Transformer encoder) and instantiate an in-season variant for each issuance month $k \in \{1, \dots, 6\}$, corresponding to September; September–October; ...; September–February.

For issuance $k$, inputs include only the first $k$ months of within-season weather. Formally, for node $g$ in municipality $m$ and crop season $t$, we define

$$X_{g,t}^{(k)} \in \mathbb{R}^{k \times P} \tag{14}$$

as the truncation of $X_{g,t} \in \mathbb{R}^{T \times P}$ to months $1{:}k$, while static covariates $(z_m, \tau_t)$ are unchanged. Each in-season model inherits the architecture, optimiser, and hyperparameters from the chosen end-of-season Transformer; only the input length differs across issuances. For every $k$, we retrain the model from scratch and evaluate it under the same LOYO protocol used for end-of-season forecasts, ensuring that all in-season metrics are based on strictly out-of-sample years.

Prediction aggregation follows the same steps as in Section 2.5, with one modification appropriate to the in-season setting. Node-level predictions are first collapsed from the spatial instance expansion representation back to the municipality level by averaging across spatial samples for a given $(m, t)$. Municipal predictions are then aggregated to national values using planted area instead of harvested area as weights, reflecting the fact that final harvested area and production are not yet observed at forecast issuance. For each issuance $k$, we compute national-scale RMSE on the LOYO test years and derive the corresponding Skill score relative to the farmer baseline (Section 2.5). In the Results, we summarise in-season performance primarily in terms of RMSE and Skill as functions of forecast issuance month. The farmer baseline, defined as a five-year rolling mean of observed national yield, is available pre-season and remains fixed within year $t$ because it depends only on $(Y_{c,t-1}, \dots, Y_{c,t-5})$. It therefore provides a realistic operational comparator for all lead times, against which the added value of the deep-learning framework at different points in the season can be directly quantified.

## 2.9 Model explainability via SHAP

To provide a transparent view of how the framework uses weather and static context, we conduct a post-hoc explainability analysis of the reference framework instantiated with a Transformer weather encoder (Section 2.2.2) using SHAP (SHapley Additive exPlanations). SHAP attributes to each input feature a local contribution to the model prediction for a given instance, grounded in cooperative game theory: the prediction is decomposed into a baseline term plus a sum of feature attributions that satisfy additivity and consistency properties (Lundberg and Lee, 2017).

Let $f(\mathbf{x})$ denote the trained Transformer's scalar output (predicted yield) for an input instance $\mathbf{x}$. For each feature $j$, SHAP approximates a Shapley value $\phi_j$ such that

$$f(\mathbf{x}) \approx f_0 + \sum_j \phi_j\,, \tag{15}$$

where $f_0$ is the expected model output under a background distribution of inputs. A positive $\phi_j$ indicates that feature $j$ increases the prediction relative to the baseline for that instance, while a negative value indicates a decreasing effect.

Because the proposed framework operates on a within-season weather sequence (with arbitrary temporal resolution) plus static covariates for all encoders, we represent each instance in a feature space that separates time-varying meteorological inputs from static context. Monthly AgERA5 variables are arranged as a set of variable–month features (e.g. "September precipitation", "December mean temperature") and concatenated with the edaphoclimatic one-hot label and the crop-year identifier. SHAP values are therefore computed with respect to a flattened feature vector that retains the interpretation of each climate variable at each month, as well as explicit attributions for edaphoclimatic region and crop year.

We employ the model-agnostic Kernel SHAP estimator, which uses a weighted linear regression to approximate Shapley values (Lundberg and Lee, 2017). For the explainability analysis, we train a reference end-of-season Transformer on the first eighteen crop seasons and hold out the last two seasons as an independent test window, mimicking a realistic deployment in which the model is fitted on historical data and applied to subsequent years. Using two held-out years rather than one also ensures that the crop-year covariate takes multiple values in the test set, which is important for assessing whether the model has learned the long-run yield trend from the year input rather than relying solely on within-season weather.

The background distribution is defined in the input space from the training years only: we construct the same feature matrix used for model fitting (weather variable–month features, edaphoclimatic label and crop-year), without including yields, and use all available training instances as background. This choice avoids additional approximation from background summarisation and ensures that the SHAP baseline reflects the full distribution of inputs seen during model fitting. As a robustness check, we also experimented with a k-means–compressed background (K = 700 representatives); global SHAP patterns and feature rankings were nearly indistinguishable from those obtained with the full background, so we report only the latter in the main analysis.

A fully LOYO-aligned SHAP analysis, in which a separate explainer would be fitted for each of the 20 LOYO models and their corresponding training sets, would require an order of magnitude more computation (on the order of hundreds of GPU-hours) and is therefore beyond the scope of this study. SHAP values are then estimated for a stratified sample of out-of-sample instances from the two held-out seasons, using the trained Transformer with fixed weights and the same preprocessing pipeline as in the main experiments.

To move from local attributions to global patterns, we aggregate SHAP values across instances. Global feature importance is computed as the mean absolute SHAP value $\mathbb{E}[|\phi_j|]$ for each feature $j$, which quantifies the average magnitude of its contribution to yield predictions. For weather variables, we summarise importance at the variable level by aggregating $|\phi_j|$ across all months associated with the same AgERA5 indicator (e.g. combining the SHAP values for "September precipitation", "October precipitation", … into a single precipitation-importance measure). We then further aggregate across all weather-related features to obtain an overall "weather importance", which can be directly compared with the importance assigned to the edaphoclimatic label and the crop-year input. Static features (edaphoclimatic label and crop-year identifier) are included in the same importance rankings, enabling a direct comparison between the influence of spatial context, long-term trend, and within-season weather.

All SHAP analyses are purely diagnostic: they do not influence model training, hyperparameter tuning or evaluation, and are conducted on the reference Transformer trained on the first eighteen crop seasons and applied to the last two seasons; they should not be interpreted as establishing causal relationships between inputs and yields.

# 3 Results and Discussion

## 3.1 End-of-season national performance

**National accuracy (end-of-season; national scale):**

Pooled across LOYO years, the Transformer delivers the lowest national RMSE at 149 kg ha$^{-1}$ (95% CI 118–203), corresponding to rRMSE 5.3% (95% CI 4.1–7.2) and $R^2$ = 0.784. Relative to the 5-year rolling-mean farmer baseline, this yields $Skill_{farm}$ = 47.6%, i.e. the Transformer's RMSE is 47.6% lower, roughly halving the typical error compared with the baseline (Table 3). Compared with ridge regression using the same inputs, the Transformer reduces RMSE by 33% and gains 26 percentage points in $Skill_{farm}$.

| Model | RMSE (kg ha⁻¹) [95% CI] | rRMSE (%) [95% CI] | Bias (kg ha⁻¹) | $R^2$ | $Skill_{farm}$ (%) |
|---|---|---|---|---|---|
| Farmer (5-yr rolling mean) | 285 [228–346] | 10.0 [8.0–11.9] | -152 | 0.216 | 0.0 |
| Ridge (linear; same inputs) | 223 [178–265] | 7.9 [6.3–9.3] | -57 | 0.518 | 21.6 |
| MLP (baseline) | 172 [129–230] | 6.0 [4.6–8.2] | -96 | 0.716 | 39.8 |
| CNN | 153 [117–214] | 5.4 [4.1–7.6] | -3 | 0.773 | 46.3 |
| LSTM | 150 [120–198] | 5.3 [4.1–7.0] | -49 | 0.783 | 47.4 |
| CNN–LSTM | 155 [116–198] | 5.4 [4.1–7.1] | -42 | 0.768 | 45.6 |
| Transformer | 149 [118–203] | 5.3 [4.1–7.2] | -8 | 0.784 | 47.6 |
| Mamba (SSM) | 156 [125–193] | 5.5 [4.4–6.8] | -43 | 0.766 | 45.4 |

Table 3: End-of-season national performance. RMSE (kg ha$^{-1}$) and rRMSE (%) are reported with 95% bias-corrected and accelerated bootstrap confidence intervals (10,000 resamples of years). Bias is the mean signed error (predicted − observed, kg ha$^{-1}$; negative values indicate under-prediction). $R^2$ is pooled across years. $\text{Skill}_{\text{farm}} = 100 * (1 - {\text{RMSE}_{\text{model}}}/{\text{RMSE}_{\text{farmer}}})$, where the farmer baseline is a 5-year rolling mean. rRMSE is normalised by the mean observed national yield. National aggregates are computed by area-weighting municipality yields using harvested area. Sensitivity to moving-block bootstrap (2–3-year blocks) is provided in Supplementary Table 2.

Across identical inputs, all deep learning variants (MLP, CNN, LSTM, CNN–LSTM, Transformer, Mamba) outperform ridge regression at the national scale (Table 3). Moreover, all sequence encoders (CNN, LSTM, CNN–LSTM, Transformer, Mamba) outperform the non-sequential MLP, consistent with advantages in modelling non-linearities and intra-season temporal structure in the monthly weather sequence.

The end-of-season results therefore indicate that sequence models deliver operationally material gains over both the five-year moving-average ("farmer") baseline and a linear ridge model using identical inputs. The Transformer's rRMSE ≈ 5.3% and $Skill_{farm}$ = 47.6% imply roughly half the error of the pre-season baseline at national scale, while a 33% RMSE reduction versus ridge (under identical covariates) underscores a substantial modelling gain beyond linear baselines.

Building on Table 3, the systematic advantage of the deep learning models over ridge is consistent with two complementary mechanisms: (i) non-linear networks can represent interaction effects and response thresholds that a linear specification cannot; and (ii) sequence encoders, in particular, exploit intra-season temporal structure (timing and accumulation effects) that an MLP without an explicit temporal inductive bias can only partially approximate. While ridge shrinkage mitigates multicollinearity, it remains linear in the features; by contrast, the sequence models learn distributed representations that reduce redundancy and combine signals across months more effectively, likely underpinning their superior national-scale accuracy.

**Model ranking:**

The top two models are the Transformer and the LSTM, and they are essentially tied at the national scale. Their pooled LOYO performance is nearly identical: RMSE 149 kg ha$^{-1}$ (95% CI 118–203) for the Transformer versus 150 kg ha$^{-1}$ (95% CI 120–198) for the LSTM; rRMSE is 5.3% for both with overlapping intervals; $R^2$ differs at the third decimal (0.784 vs 0.783); and MAE differs by 3 kg $ha^{-1}$ (124 vs 127). A paired Wilcoxon signed-rank test on year-level absolute errors found no detectable systematic advantage ($W = 104$, $p = 0.985$, $n = 20$). We therefore use the Transformer for downstream analyses on pragmatic grounds: in our implementation it trains about three times faster and its attention mechanism is expected to scale more naturally to longer input sequences (for example weekly or daily weather), which we explore in follow-up work. Beyond training speed, the Transformer's inductive bias is also well aligned with this setting. Self-attention can integrate signals from any month-to-month pairing, which is useful when yield anomalies are driven by distributed within-season patterns (e.g., cumulative moisture conditions followed by late-season heat or radiation constraints) rather than strictly local or strictly long-range dependencies. The learned positional embedding preserves seasonal ordering, while attention allows the model to re-weight informative periods without assuming a fixed receptive field. Together, these properties plausibly help explain why the Transformer performs strongly under identical inputs in the national LOYO benchmark. Where near-ties exist in future applications, deployment choices can weigh accuracy against computational footprint and maintainability; criteria such as inference speed, hardware availability, and ease of monitoring may drive selection without sacrificing national-scale accuracy.

**State-level accuracy:**

To characterise sub-national error structure and assess whether national performance masks systematic state-level misfit, we aggregated municipal predictions to state × year using harvested-area weights and computed per-state errors pooled across LOYO years ($n = 20$ per state). State-level performance for the reference Transformer is reported in Table 4. Several major producing states, which together accounting for about 70% of total soybean production in the study domain (Figure 3a), show comparatively small mean biases (MT −19, PR −3, GO −2, and RS 16 kg $ha^{-1}$), and the pooled national bias remains near zero (−8 kg $ha^{-1}$; Table 3). At the same time, error magnitudes remain heterogeneous across states (RMSE 145–535 kg $ha^{-1}$; MAE 107–385 kg $ha^{-1}$), and some states exhibit non-negligible signed biases, including larger negative bias in MS (−115 kg $ha^{-1}$) and MG (−76 kg $ha^{-1}$), and positive bias in BA (88 kg $ha^{-1}$) and PI (84 kg $ha^{-1}$). These state-level diagnostics therefore indicate that sub-national residual structure exists and should not be overlooked, while suggesting that national skill is not explained solely by a simple cancellation of large opposing state-scale biases. Stronger evidence on this point is provided by the municipality-level generalisation diagnostics reported next.

| State | RMSE (kg ha⁻¹) | MAE (kg ha⁻¹) | Bias (kg ha⁻¹) | Years |
|---|---|---|---|---|
| BA | 455 | 366 | 88 | 20 |
| GO | 180 | 134 | -2 | 20 |
| MA | 279 | 191 | 52 | 20 |
| MG | 203 | 158 | -76 | 20 |
| MS | 320 | 253 | -115 | 20 |
| MT | 145 | 107 | -19 | 20 |
| PI | 535 | 385 | 84 | 20 |
| PR | 279 | 245 | -3 | 20 |
| RS | 377 | 287 | 16 | 20 |
| TO | 164 | 135 | 30 | 20 |

Table 4: State-level performance. Per-state errors pooled across LOYO years (n = 20 per state). Municipal predictions were aggregated to state × year using harvested-area weights; metrics are computed on the resulting yearly state series. RMSE and MAE are in kg $ha^{-1}$. Bias is mean signed error (predicted − observed, kg $ha^{-1}$).

### Municipality-level generalisation diagnostic

To strengthen this assessment at a finer spatial scale, we next examine performance at the municipality level. For each municipality, we compute the Pearson correlation between observed and predicted yields across LOYO years. The resulting map (Figure 4) shows predominantly positive correlations across the evaluated soybean-producing municipalities, with moderate-to-high values across large contiguous areas rather than a patchwork of sharply contrasting positive and negative correlations. This spatial coherence indicates that the model captures interannual variability in a geographically consistent way at the municipal level. While correlation does not measure error magnitude or bias, it provides a direct diagnostic that national performance does not arise primarily from geographically inconsistent local tracking.

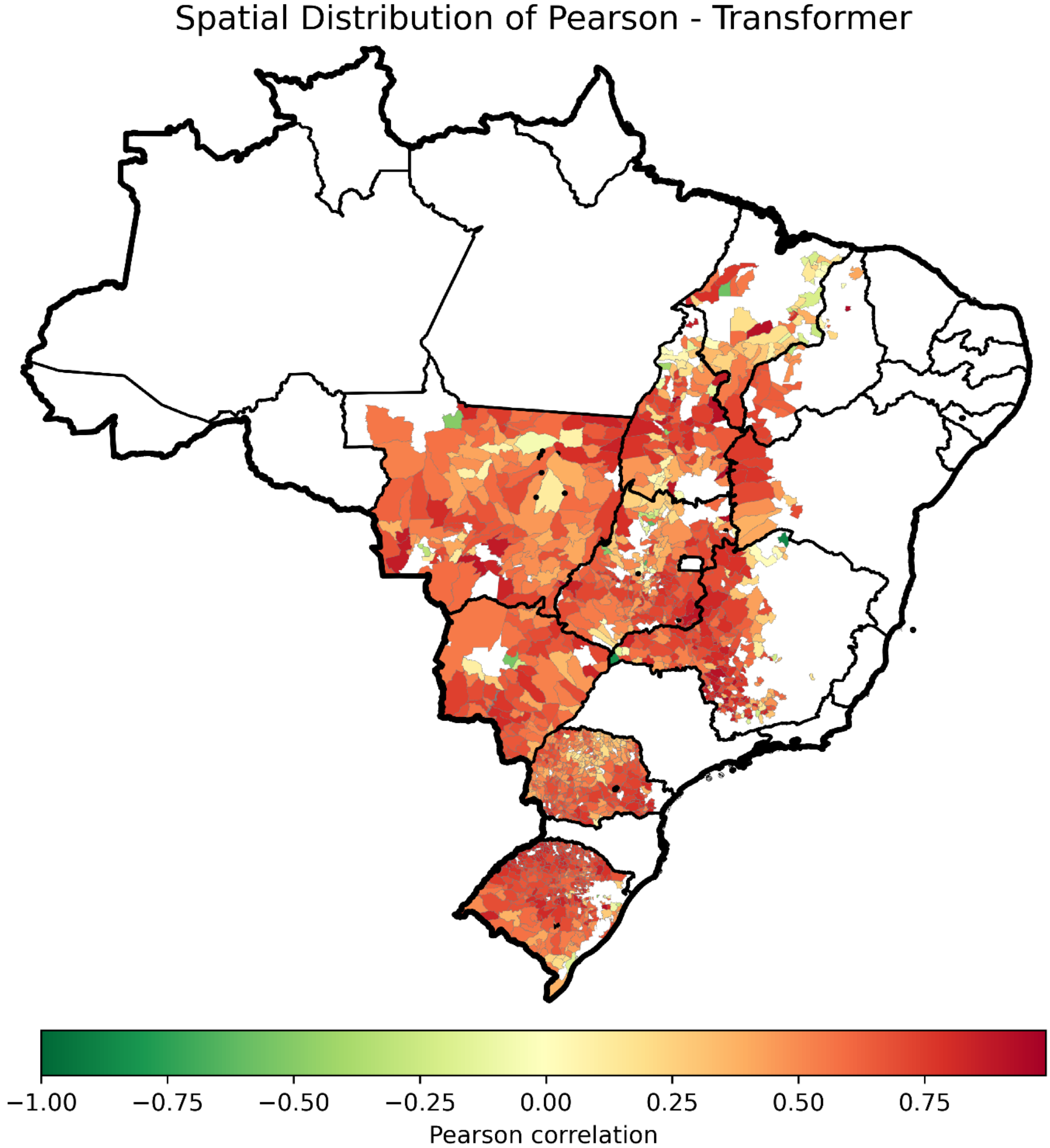

Figure 4: Municipality-level agreement map (Transformer model). Spatial distribution of Pearson correlations between observed and predicted municipality-level yields across LOYO years (2001–2020). Predominantly positive correlations across the soybean-producing region indicate that the model captures the direction and timing of interannual variability at the municipal scale. Values closer to zero indicate weaker temporal tracking; negative values indicate inversions in year-to-year variation.

To quantify the prevalence and strength of temporal tracking, we summarise the distribution of municipality-level correlations (Figure 5). Across 1,487 municipalities, the median correlation was $r = 0.629$ with an interquartile range of 0.465–0.734. Correlations were positive for 96.7% of municipalities, and only 3.3% exhibited negative values. Moreover, 87.4% of municipalities exceeded $r \geq 0.30$ and 70.5% exceeded $r \geq 0.50$, while near-zero correlations were uncommon ($|r| < 0.10$ in 2.4% of municipalities). Here, we use $r \geq 0.30$ and $r \geq 0.50$ as conventional “rule-of-thumb” thresholds for moderate and large/strong associations, respectively, consistent with widely used effect-size guidelines for Pearson's r (Cohen, 2013). Together, these distributional summaries reinforce the map-based interpretation that the model's national performance reflects broadly consistent municipality-level tracking of year-to-year yield variability, rather than spatial averaging over broadly conflicting municipality-level dynamics.

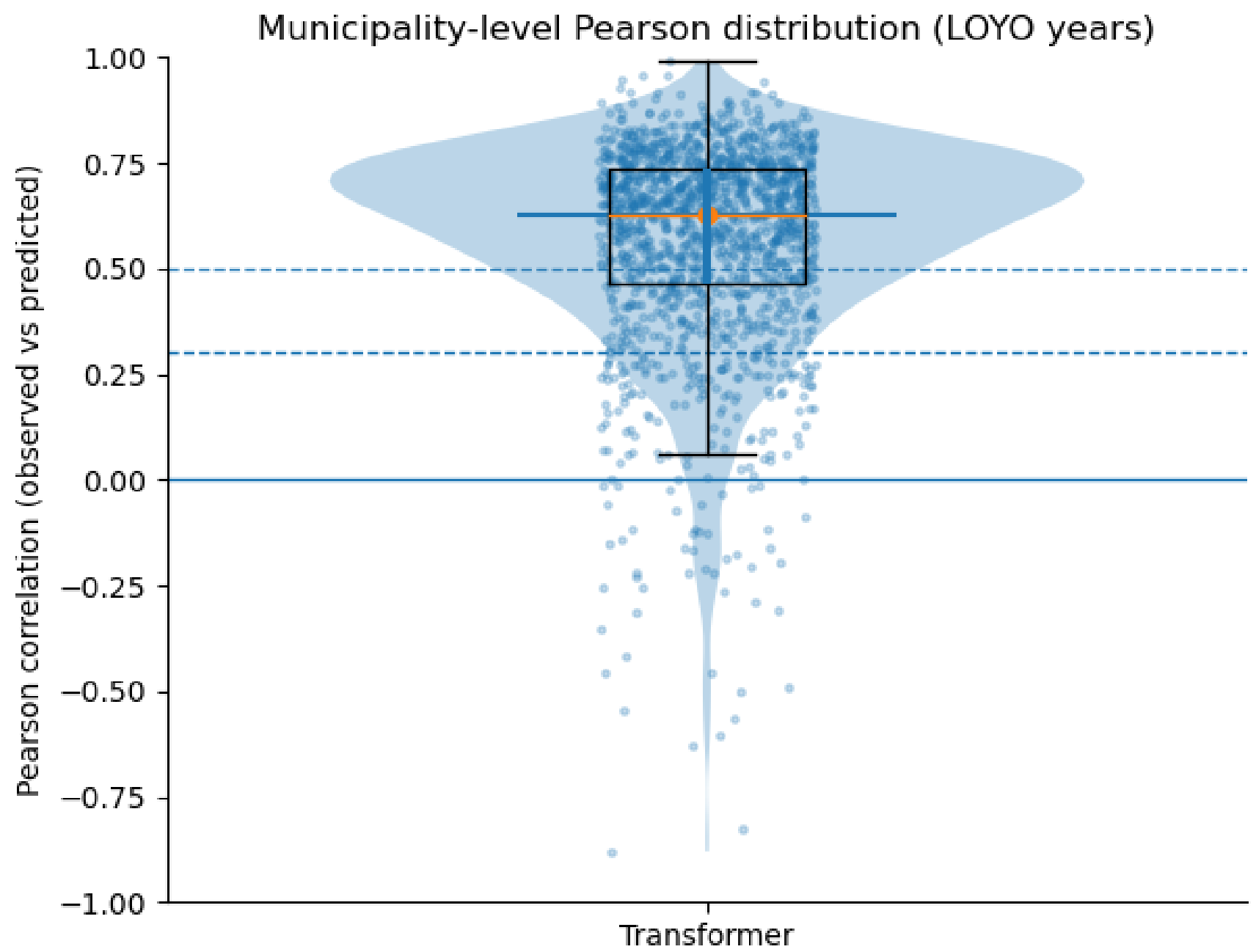

Figure 5: Municipality-level Pearson distribution (Transformer model). Violin plot (with overlaid boxplot and jittered municipality points) summarising Pearson correlations between observed and predicted municipality-level yields across years (2001–2020). The violin width reflects the density of municipalities at each correlation value, and the boxplot indicates the median and interquartile range (IQR). Predominantly positive correlations are observed (median $r = 0.629$, IQR 0.465–0.734; 96.7% of municipalities with $r > 0$), with a large majority exceeding conventional thresholds (87.4% with $r \geq 0.30$, 70.5% with $r \geq 0.50$; dashed lines), indicating consistent temporal tracking of interannual yield variability at the municipal scale.

**National time series and observed–predicted agreement.**

For the Transformer model, observed and predicted national-yield time series show similar interannual variation (Figure 6). The model captures the decline and recovery in the early 2000s (2001–06), the high around 2010–11, the drop in 2012, and the subsequent rise through 2017–19. Deviations are moderate (for example, an under-prediction near the 2011 peak and an over-prediction in 2012) and there is no clear temporal drift.

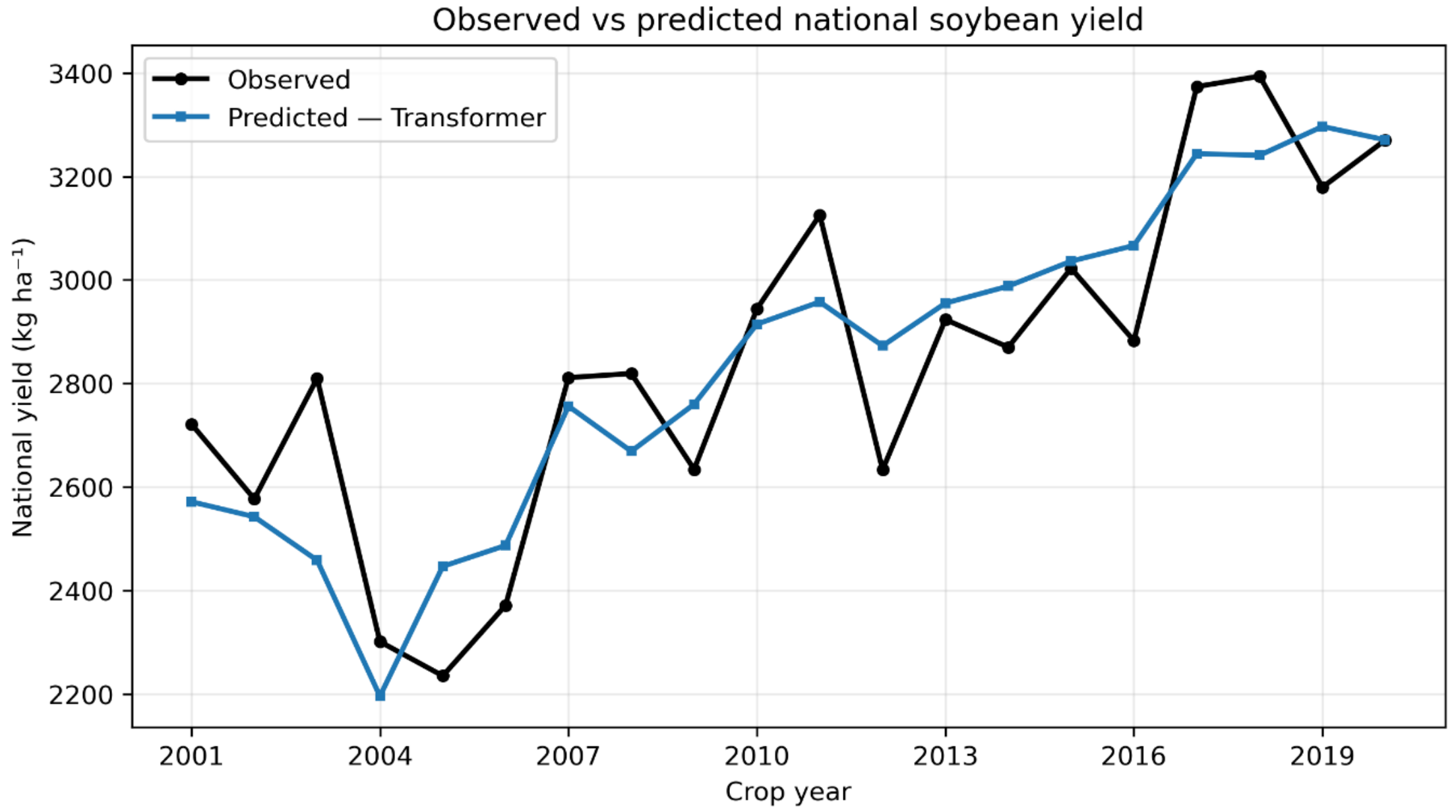

Figure 6: Time series of national yield (observed vs predicted, Transformer model)

To test whether the national agreement in Figure 6 could arise from spatial aggregation masking regional misfit, we next examine observed–predicted trajectories at the state scale (Figure 7). The state panels provide a direct visual check for systematic temporal drift and for persistent state-specific under- or over-prediction that could cancel out at the national level. Consistent with the pooled state diagnostics (Figure 6), the Transformer generally reproduces the timing and direction of interannual yield changes across major producing states, with discrepancies that are episodic rather than monotonic over time. Taken together, the state-level trajectories support the interpretation that national skill is not primarily an artefact of offsetting regional biases, although heterogeneous departures in individual years and smaller producers indicate that some residual sub-national error structure remains.

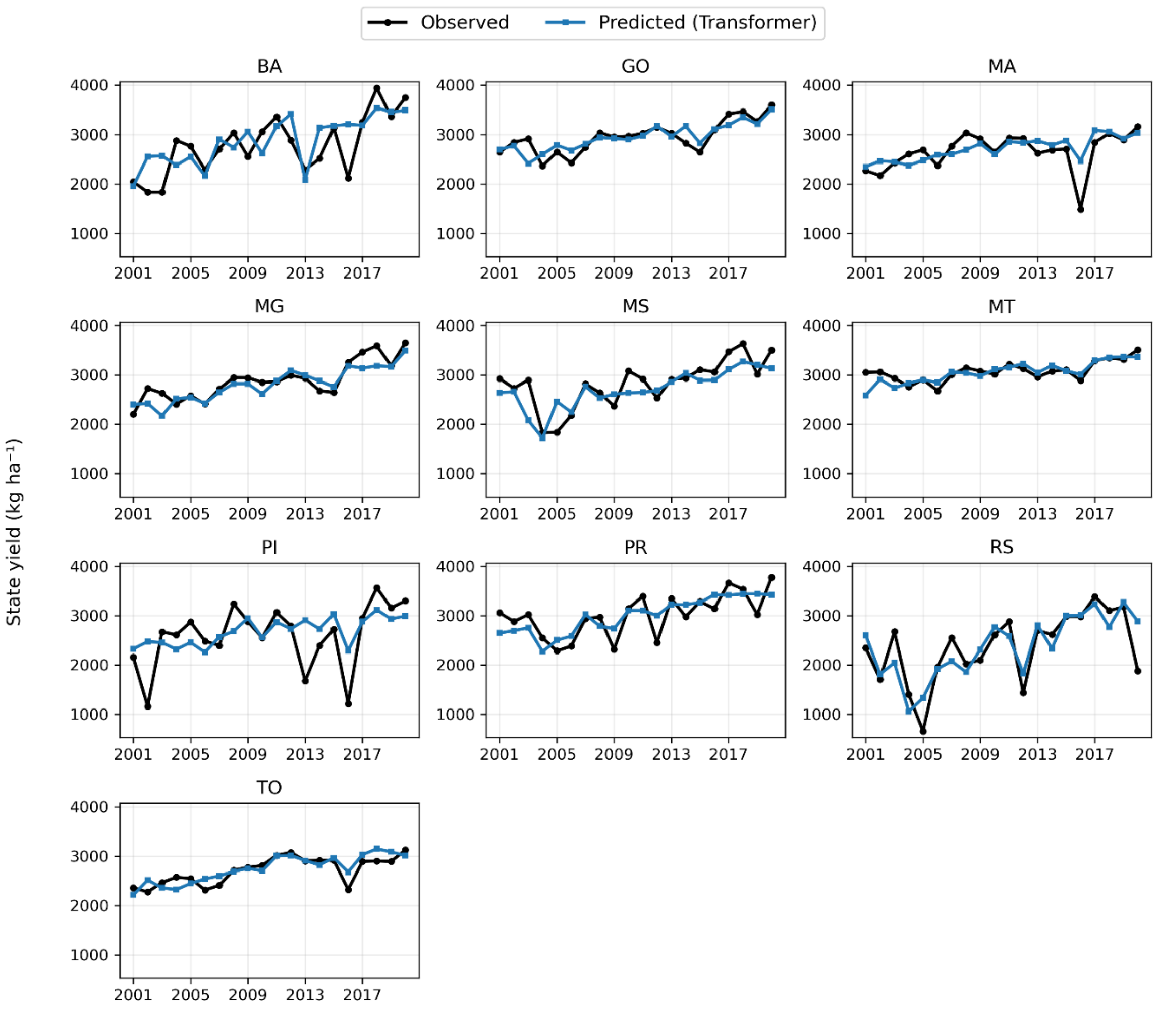

Figure 7: State-level observed vs predicted soybean yield time series (Transformer model). Observed and predicted soybean yield time series at the state scale for the reference Transformer model. Municipal predictions were aggregated to state × year using harvested-area weights, and yields are shown in kg $ha^{-1}$ for the 2001–2020 crop years. Panels share a common y-axis to facilitate cross-state comparison of both yield levels and interannual variability. Across states, the model broadly tracks interannual fluctuations without evidence of systematic temporal drift, while state-specific departures highlight where residual errors remain concentrated at sub-national scales.

The scatter of observed versus predicted national yields (Figure 8) clusters around the 1:1 line, with $R^2$ = 0.784, RMSE = 149 kg $ha^{-1}$, MAE = 124 kg $ha^{-1}$ and mean error −8 kg $ha^{-1}$ (i.e. negligible bias). Dispersion about the diagonal is broadly uniform across the yield range, with no evident fan-shape.

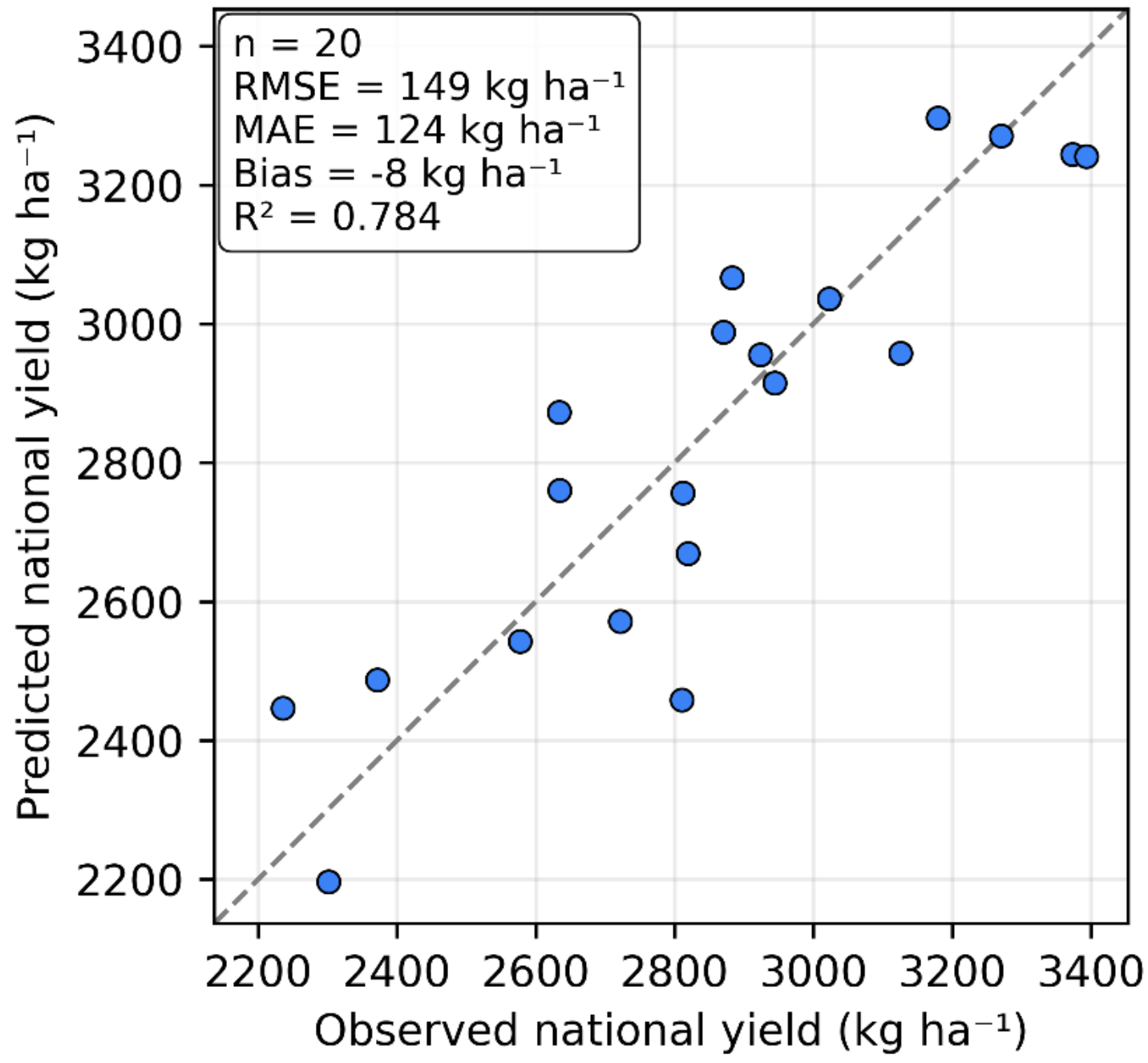

Figure 8: Observed vs predicted national soybean yield — Transformer model (Brazil, 2001–2020). Scatter of observed national yield against transformer predictions for 20 crop years (2001–2020). Municipality-level forecasts were aggregated to country level using harvested-area weights. The dashed line is 1:1.

Together, the alignment with the 1:1 line, the roughly uniform scatter, and the near-zero national bias indicate good calibration in the sense of agreement between predictions and observations at the aggregate level (Wilks, 2011). In addition, the municipality-level correlation map and its distributional summary show predominantly positive associations across the soybean-producing region, indicating geographically coherent tracking of interannual variability beyond the national aggregate. Taken together, these diagnostics suggest that the Transformer model captures both the long-term trend and year-to-year variability at the national scale, while state-level and municipality-level analyses confirm that residual sub-national structure remains but does not, by itself, account for the observed national skill.

## 3.2 Contributions of edaphoclimatic label and spatial instance expansion

This subsection quantifies the contribution of two independent components of the proposed methodology: the edaphoclimatic region label and spatial instance expansion.

**Edaphoclimatic label:**

Removing the edaphoclimatic label increased national RMSE from 149 to 162 kg ha$^{-1}$ (+8.7%) and reduced $Skill_{farm}$ from 47.6 to 43.0 percentage points (Table 5). To assess whether this effect also holds at the operational (municipality) scale, and to account for (i) heteroskedasticity across municipalities with different harvested areas and (ii) year-to-year dependence driven by common shocks, we analysed area-weighted municipality–year absolute errors using weighted least squares with cluster-robust standard errors by year (20 clusters). Under this specification, removing the edaphoclimatic label increased municipality-level absolute error by 20.5 kg ha$^{-1}$, and this effect was statistically significant (p = 0.0158; 95% CI 2.7–38.3; n = 26,063; cluster-robust by year).

| Setting | RMSE (kg ha$^{-1}$) | rRMSE (%) | $R^2$ | $Skill_{farm}$ (%) |
| --- | --- | --- | --- | --- |
| Transformer with edaphoclimatic label | 149 | 5.3 | 0.784 | 47.6 |
| Transformer without edaphoclimatic label | 162 | 5.7 | 0.745 | 43.0 |

Table 5: Effect of edaphoclimatic region

These results are consistent with the interpretation that soil–climate context, as summarised by the edaphoclimatic label, improves the mapping from weather to yield by capturing persistent regional differences in production environments. Even though the label is a coarse proxy for soils, management and cultivar choices, its inclusion yields a measurable and practically relevant improvement in accuracy at both national and municipal scales.

**Spatial instance expansion:**

Disabling SIE increased national RMSE from 149 to 165 kg ha$^{-1}$ (+10.7%) and reduced $Skill_{farm}$ from 47.6 to 42.1 percentage points (Table 6).

| Setting | RMSE (kg ha$^{-1}$) | rRMSE (%) | $R^2$ | $Skill_{farm}$ (%) |
| --- | --- | --- | --- | --- |
| Transformer w/ SIE | 149 | 5.3 | 0.784 | 47.6 |
| Transformer without SIE | 165 | 5.8 | 0.737 | 42.1 |

Table 6: Effect of spatial instance expansion

The framework is intended as a methodological template that transfers across crops and regions. While Brazilian soybean offers abundant historical data, other contexts may have sparser records (for example short cultivation histories, smaller planted areas, or gaps in official statistics). To emulate such limited-data regimes, we retained the full set of training years in each LOYO fold but randomly reduced the number of training instances within those years to 25%, 50% and 75% of the available data, respectively, and then compared annual country-level absolute errors for SIE-disabled minus SIE-enabled models ($\Delta = | E |_{no\text{-}SIE} - | E |_{SIE}$) using a one-sided Wilcoxon signed-rank test (alternative $H_1$: $\Delta > 0$, i.e. SIE reduces error). The resulting p-values are summarised in Table 7. The 50% case trends positive (SIE-favoured) but remains non-significant at $\alpha = 0.05$

| Training data fraction (% of instances) | Paired LOYO years (n) | p-value (one-sided Wilcoxon, $H_1$: $\Delta > 0$) |
|---|---|---|
| 25% | 20 | 0.392 |
| 50% | 20 | 0.066 |
| 75% | 20 | 0.816 |

Table 7: Effect of spatial instance expansion under reduced-training regimes, based on one-sided Wilcoxon signed-rank tests on national absolute errors ($\Delta = |E|_{no-SIE} - |E|_{SIE}$) across LOYO years. The alternative hypothesis is $H_1$: $\Delta > 0$ (SIE reduces absolute error).

Taken together, the ablations of the agro-environmental context label and spatial instance expansion (SIE) indicate that both components contribute to accuracy. The agro-environmental context label yields a statistically significant reduction in municipality-level absolute error (~20 kg ha$^{-1}$ on average), consistent with the view that conditioning on agro-environmental context improves predictive mapping. Spatial instance expansion does not degrade performance on average and shows directional benefit when training data are scarce (notably at 50% of training years). Although reduced-data effects are not significant at $\alpha = 0.05$, the combination of (i) higher full-data accuracy with SIE and (ii) no evidence of harm under sub-sampling supports its use as a robustness device. These results justify including both the agro-environmental context label and SIE in the proposed template for transfer to other crops and regions, particularly where historical records are limited.

## 3.3 In-season performance and lead time

We assess forecast lead-time behaviour by evaluating an in-season version of the reference framework instantiated with the Transformer weather encoder, using progressively truncated within-season inputs (Section 2.8). Each issuance month $k$ uses only the first $k$ time steps of the within-season weather sequence (Equation 14), while static context inputs are unchanged. All in-season models are trained and evaluated under the same LOYO protocol as the end-of-season setting, and national-scale performance is summarised using RMSE and $Skill_{farm}$ relative to the pre-season farmer baseline (Sections 2.5 and 2.8).

To reflect operational conditions at issuance, national aggregation uses planted-area weights (rather than harvested area), and node-level outputs are collapsed back to municipality predictions consistent with the spatial instance expansion scheme (Sections 2.1.3 and 2.5). We therefore report RMSE and $Skill_{farm}$ as functions of issuance month, allowing a direct readout of how quickly the framework adds value over the baseline as the season unfolds.

Figure 9 shows progressive improvement as within-season information accumulates. National RMSE declines from 251 kg ha$^{-1}$ (September) to 146 kg ha$^{-1}$ (February), with intermediate values 278, 219, 220, and 173 kg ha$^{-1}$ for October–January, respectively. The corresponding $Skill_{farm}$ rises from 12.6% (September) to 49.0% (February), passing through 3.2%, 23.7%, 23.3% and 39.6% for October–January. The largest month-on-month gain occurs between December and January (RMSE 220 → 173; $Skill_{farm}$ 23.3% → 39.6%).

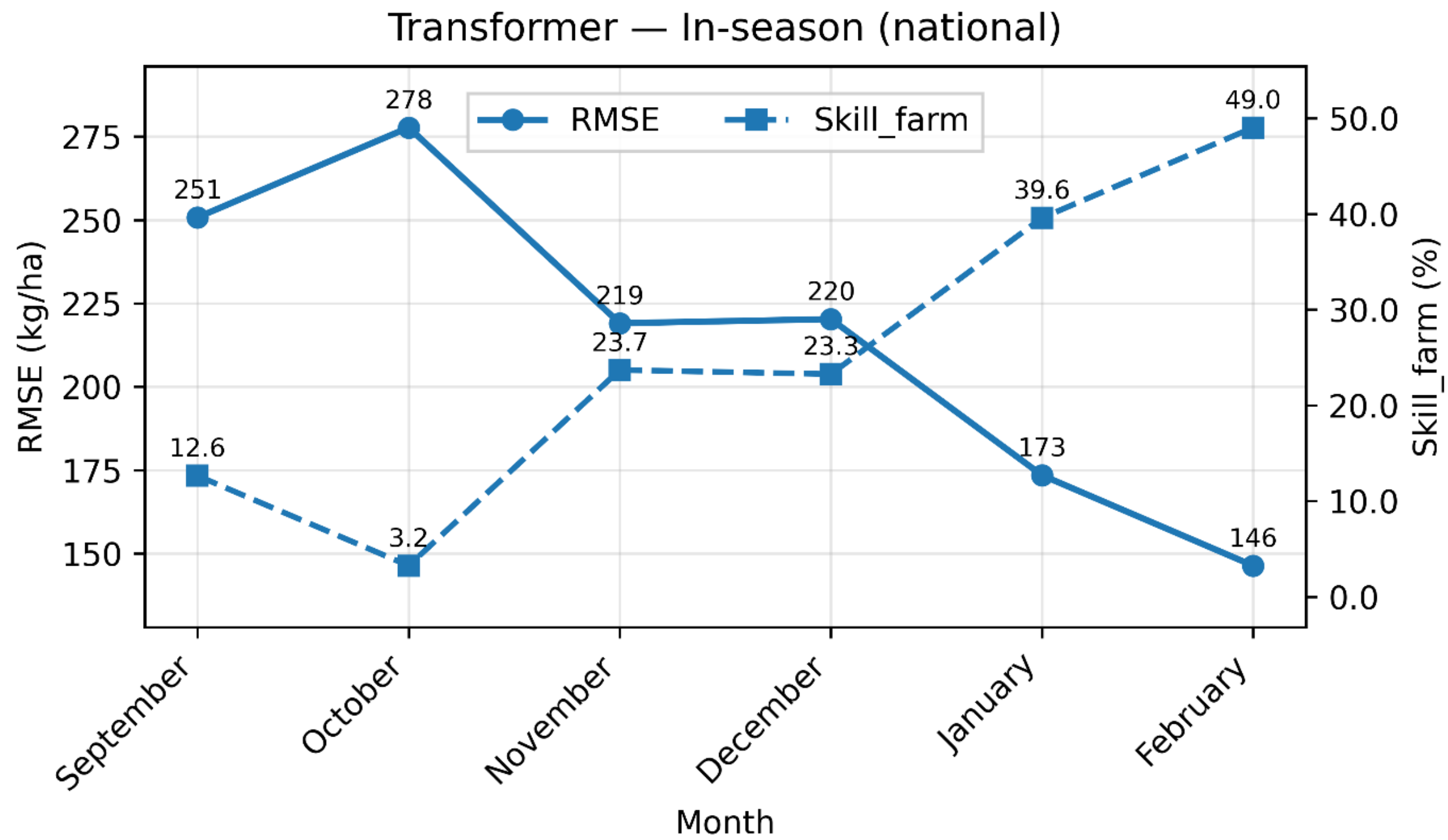

Figure 9: Transformer model in-season performance at national scale (Brazil). RMSE (left axis, kg ha$^{-1}$) and $Skill_{farm}$ (right axis, %) by cut-off month. Forecasts are aggregated from municipality to country using planted-area weights and evaluated over 20 crop years (2001–2020). The farmer baseline relies only on the previous five harvests and is invariant across months.

Building on this in-season evaluation, the Transformer surpasses the baseline at every cut-off. Even early in the season (September–November) $Skill_{farm} > 0$, indicating consistently lower error than the baseline on these data and thereby offering earlier, more reliable guidance for procurement, logistics and risk-management decisions. This is operationally relevant because it implies that meaningful reductions in forecast error can be achieved months before harvest, when many market and supply-chain decisions are still adjustable.

## 3.4 Positioning relative to previous yield-forecasting studies

Our end-of-season national performance (rRMSE 5.3% [4.1–7.2]; RMSE 149 kg ha$^{-1}$ [118–203]) is competitive with Brazil benchmarks that often rely on substantially richer, higher-cost input pipelines than our weather + edaphoclimatic label feature set. For context, Table 8 summarises the key Brazil-focused soybean-yield studies considered here, their input modalities, spatial scope, and the headline accuracy metrics reported.

For example, Bloh, von et al. (2023) combine satellite processing (cropland masking and municipality-averaged pixels) and multiple remote-sensing indices (NDVI, EVI, CVI, GLI) with engineered agro-climatic stress features and an ENSO index; their feature engineering explicitly encodes agronomic prior knowledge around heat and drought constraints. They further compute national yield via production-share weighting across municipalities and report national rRMSE as low as 4.8% for their ensemble model.

Similarly, Song et al. (2022) build a Brazil-wide yield mapping pipeline that aggregates multi-source covariates to municipal scale, including remote sensing, topographic features, climate and weather variables, and soil properties, followed by bias correction and model composition. Their reported accuracies (e.g., RMSE of 344 kg $ha^{-1}$) arise from a data-rich workflow designed for high-resolution mapping rather than a frugal national forecasting setting.

Monteiro et al., (2022) highlight that adding remote-sensing products (e.g., Landsat/Sentinel) could improve accuracy, indicating that their evaluated models are not exploiting the full satellite feature space available in related work; their reported RMSE levels for RF/SVM commonly fall around 400–500 kg $ha^{-1}$ across validation resampling scenarios.

Finally, Cunha, Silva and Netto (2018) propose a multi-source pre-season system explicitly incorporating satellite-derived precipitation, soil properties, and seasonal climate forecast data (motivated partly by the costs and timing constraints of NDVI-based approaches), reporting RMSE 386 kg $ha^{-1}$.

Taken together, these comparisons suggest that our model attains state-of-the-art national end-of-season accuracy in Brazil with materially simpler inputs, supporting the claim that a frugal, operationally plausible feature set (weather + edaphoclimatic label) can reach the same performance neighbourhood as more complex pipelines that incorporate remote sensing, engineered stress indices, and/or soil datasets.

Because input data, spatial unit, and validation strategies differ across studies, these benchmarks are interpreted as contextual rather than strictly like-for-like. Moreover, most prior studies do not report performance relative to a simple, fully replicable operational benchmark (such as the five-year “farmer” baseline used here). This limits the feasibility of benchmark-normalised comparisons that could partially adjust for differences in period, spatial coverage, and yield variability across studies (Filippidis , M ., Filis , G . and Magkonis; 2024), and reinforces that Table 8 is intended for positioning rather than strict cross-study ranking.

| Study | Horizon used for comparison | Inputs and pipeline complexity (high-level) | Validation design (as reported) | Reported performance |
|---|---|---|---|---|
| This work (Transformer model) | End-of-season | Frugal: weather predictors + edaphoclimatic label (no remote sensing, no soil layers, minimal feature engineering) | LOYO, 2001–2020 (aligned window to von Bloh) | rRMSE 5.3% [4.1–7.2]; RMSE 149 kg $ha^{-1}$ [118–203] |
| Bloh, von *et al.* (2023) | End-of-season | Data-rich: satellite processing + vegetation indices (e.g., NDVI/EVI) and engineered agro-climatic stress features informed by agronomic prior knowledge, plus climate and ENSO context | LOYO, 2001–2020 | National: rRMSE 4.8% (Ensemble), 4.9% (ANN); baseline LR rRMSE 6.7%, RMSE 0.2 t $ha^{-1}$ |
| Monteiro *et al.*, (2022) | End-of-cycle | Weather-based setup evaluated across multiple scenarios; authors note remote sensing products (e.g., Landsat/Sentinel) as potential additions for improvement | Repeated random splits / resampling (end-of-cycle summarised as distributions rather than a single LOYO national metric) | For 60/90/120 DAS, RF/SVM RMSE typically ~400–500 kg $ha^{-1}$ (≈12–15%) |
| Song *et al.* (2022) | End-of-season / post-season mapping context | Very data-rich: multi-source covariates including remote sensing, topography, climate/weather, and soil properties, plus bias correction | Withheld test split + LOYO analysis reported | Overall RMSE 344 kg $ha^{-1}$, $R^2$ 0.69 (~12% relative error); LOYO RMSE range 259–816 kg $ha^{-1}$ |
| Cunha, Silva and Netto (2018) | Pre-season / system context (not end-of-season) | Multi-source system: includes satellite-derived precipitation, soil properties, and seasonal climate forecast information (explicitly motivated by timing/cost constraints of NDVI-heavy workflows) | Not fully extractable from the single benchmark row alone | For "Brazil Soybean": RMSE 385.81, $R^2$ 0.55, RMSPE 14.31 |

Table 8: Contextual benchmarking of Brazil soybean-yield modelling studies (end-of-season focus). This table provides a targeted, contextual benchmark for Brazil soybean-yield modelling studies reporting RMSE and/or relative error metrics. The primary like-for-like reference is von Bloh et al. (2023), which matches national aggregation, the 2001–2020 window, LOYO evaluation, and defines rRMSE as RMSE normalised by mean observed yield. Other studies are included as Brazil-wide context but differ in target scale, horizon timing, and/or validation protocol; therefore, comparisons are interpreted as indicative rather than strictly equivalent.

## 3.5 SHAP-based interpretation of the reference Transformer

Figure 10 summarises how the reference Transformer distributes predictive importance across the three feature groups (weather, edaphoclimatic region and crop year), based on mean absolute SHAP values aggregated over all municipal instances in the two held-out seasons. At this grouping level, the crop-year input clearly dominates: it accounts for roughly three-quarters of the total importance, with a mean | SHAP |of 760 kg ha$^{-1}$ compared with 197 kg ha$^{-1}$ for weather and 94 kg ha$^{-1}$ for edaphoclimatic region. In relative terms, year carries about four times as much importance as weather and about eight times as much as the edaphoclimatic label, indicating that the model relies heavily on the crop-year covariate to anchor long-run yield levels, with weather and regional context providing secondary but non-negligible adjustments around this baseline.

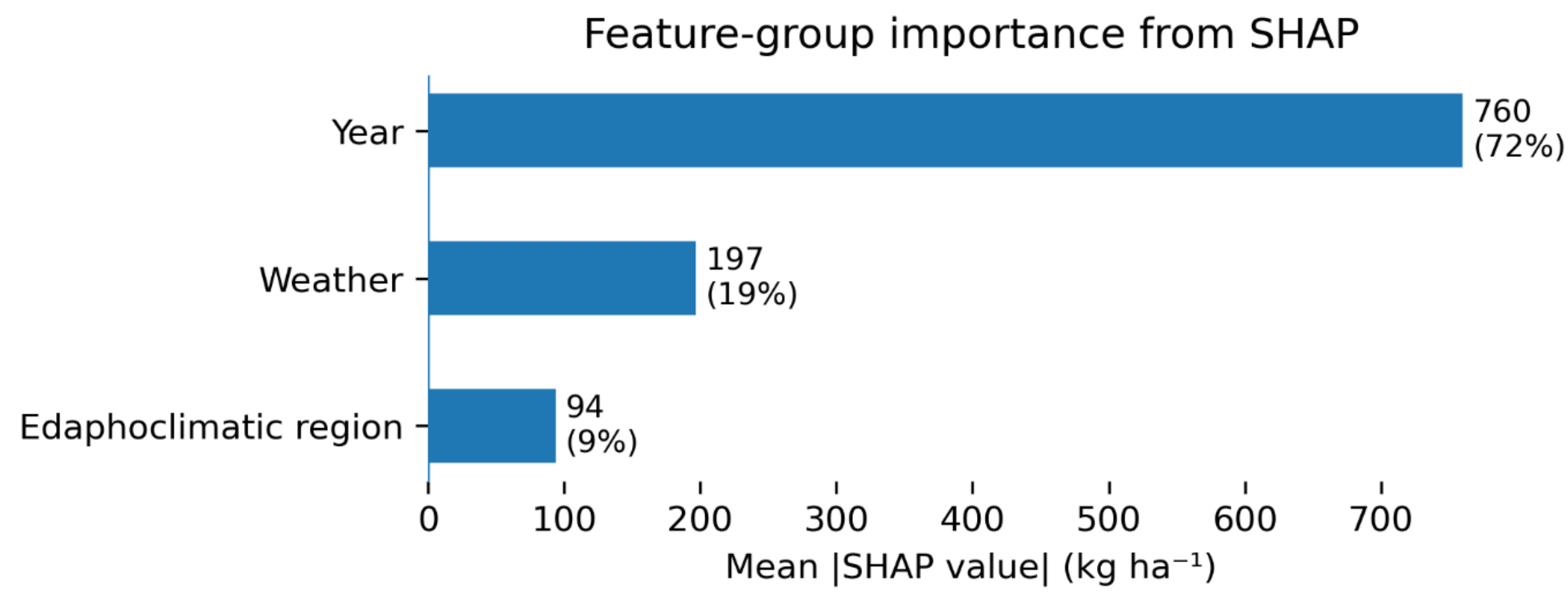

Figure 10: SHAP feature-group importance for the reference Transformer model. Bars show the mean absolute SHAP value for each feature group (weather variables aggregated over all months, edaphoclimatic region and crop year), computed across all municipal instances in the two held-out seasons. Higher values indicate a larger average contribution of that group to the model's yield predictions.

This decomposition is consistent with the strong upward trend in Brazilian soybean yields over 2001–2020 (Figure 6) and with our decision to provide the crop year as an explicit input instead of detrending yields *a priori*. The SHAP results indicate that the Transformer has indeed assigned most of the structural, long-horizon variation to the year feature, while using weather and edaphoclimatic region primarily to explain shorter-run deviations around that trend. Although their aggregate importance is smaller than that of year, both weather and regional context still contribute meaningfully to predictions, supporting the view that high-frequency meteorological variability and persistent spatial heterogeneity are necessary to refine forecasts beyond what could be obtained from technological progress alone. These attributions should be interpreted as model-based associations rather than causal effects, but they provide reassurance that the network is using the crop-year covariate in the intended way and not simply memorising idiosyncratic noise in the yield series.

Figure 11 shows the SHAP beeswarm plot for the crop-year input using the two held-out seasons at the end of the study period. All SHAP contributions are positive, indicating that, relative to the model's average prediction over the training-set background, the year feature systematically increases predicted yields in these recent seasons. Points corresponding to higher values of year (later crop year; red) are concentrated at larger SHAP values than those for lower year (earlier crop year; blue), meaning that the model assigns a stronger positive adjustment to the most recent year. This pattern is consistent with the observed upward trend in Brazilian soybean yields over the study period and shows that the network has encoded this secular increase via the year input, over and above the effects captured by weather and edaphoclimatic variables. From an agronomic perspective, this is plausible: each new crop year tends to bundle gradual improvements in genetics, management practices and technology adoption, and the year feature acts as a coarse proxy for these cumulative gains.

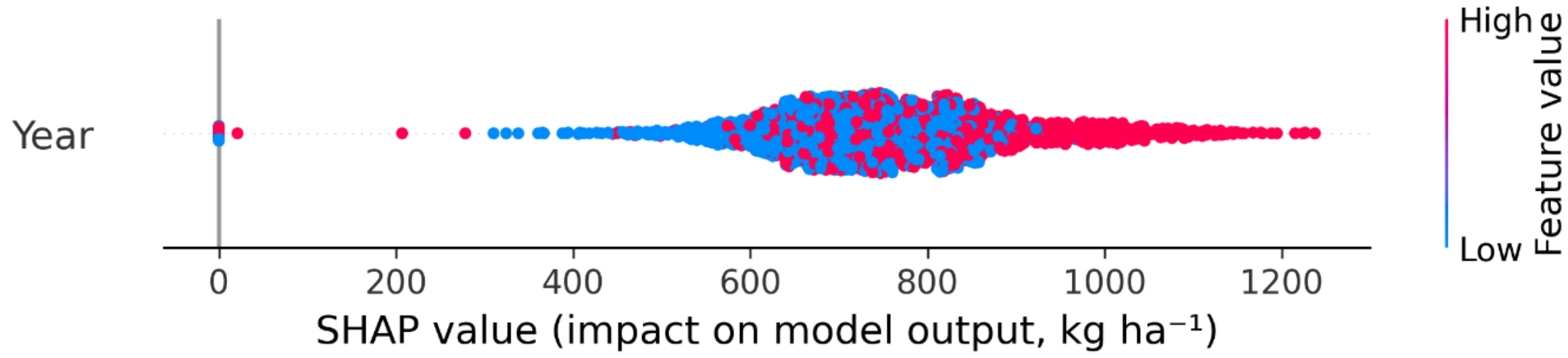

Figure 11: SHAP beeswarm for the crop-year input. Beeswarm plot of SHAP values for the crop-year feature for all municipal instances in the two held-out seasons. Each point represents one prediction; colour encodes the feature value (blue = earlier crop year, red = later crop year). All SHAP values are positive, and higher year values cluster at larger SHAP magnitudes, indicating that the model systematically increases predicted yields for more recent years, consistent with the observed long-run technological and management-driven yield gains.

The strong SHAP importance of the year input highlights that a substantial share of predictive power comes from the historical yield trajectory itself. Rather than separating trend handling into explicit detrending and post-hoc retrending steps, as is common in the literature (Li *et al.*, 2023; von Bloh *et al.*, 2023), we deliberately incorporated crop year directly as a covariate and allowed the deep-learning encoder to learn the long-run trajectory jointly with weather-driven deviations. This design simplifies pre- and post-processing and favours practical deployment, while also providing a transparent decomposition between trend and seasonal signal. As with any trend-aware forecasting approach, however, if the underlying technological trajectory or management baseline shifts materially, the model must be updated as new crop years become available so that the learned relationship remains aligned with the evolving system.

Figure 12 shows the global SHAP importance of each weather variable, aggregated over months, together with a clustering of features by redundancy (Lundberg, 2018). Cloud fraction emerges as the single most influential weather predictor, followed by variables related to precipitation and humidity. The dendrogram reveals a first cluster linking cloud fraction, precipitation duration fraction, relative humidity, and precipitation rate, indicating that the model uses these moisture- and cloud-related variables in a partly redundant way. A second cluster groups vapour pressure, air temperature, and vapour-pressure deficit, consistent with their shared role in representing thermal conditions and atmospheric demand on the crop. Solar flux and wind speed also display substantial mean |SHAP| values but are less tightly clustered, suggesting additional, partly independent information related to radiation and boundary-layer dynamics. Solid precipitation has negligible importance, which is plausible given that events such as hailstorms, while potentially severe at field scale, are relatively rare and spatially localised within Brazilian soybean regions; at the municipal-and-national aggregation used here, their signal is diluted and does not emerge as a dominant predictor in the last two seasons. Because many meteorological predictors are physically coupled, SHAP cannot uniquely disentangle their individual effects: importance is naturally shared across collinear variables within each cluster. In this sense, the clusters are better interpreted as bundles of related processes (moisture regime, thermal conditions, radiation, and wind environment) than as strictly independent drivers. Overall, this structure aligns with agronomic understanding that sunlight, water supply and temperature are key controls on soybean yield (Farias, Nepomuceno and Neumaier, 2007), while the SHAP-based clusters should be interpreted as overlapping signals in how the model uses these variables rather than as proof of direct causal relationships.

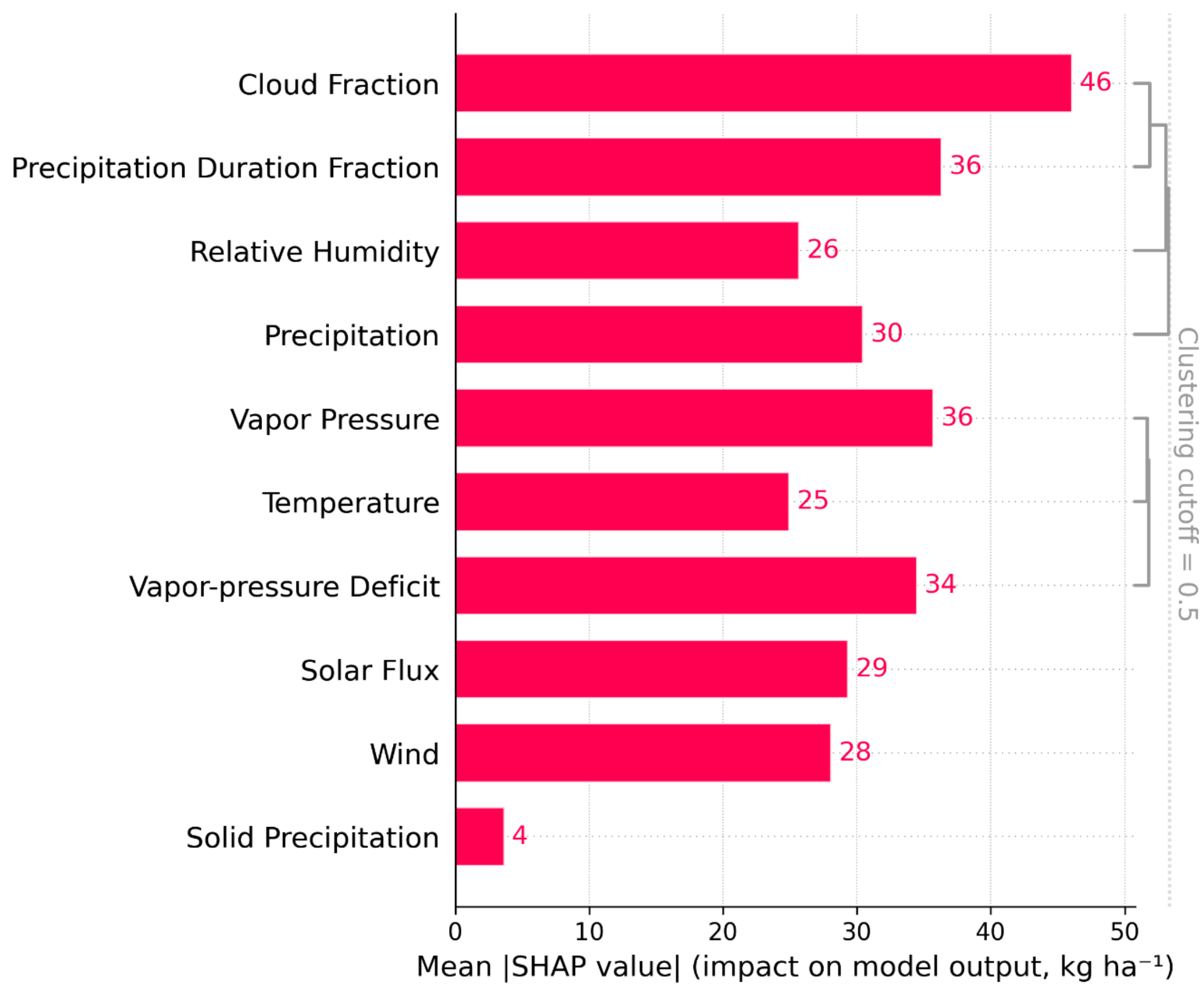

Figure 12: Global SHAP importance and redundancy structure of weather variables. Mean absolute SHAP value for each AgERA5 weather variable, aggregated over months, for the reference Transformer model on the two held-out seasons. Bars indicate global importance, and the dendrogram groups variables with similar SHAP patterns (clustering cutoff = 0.5). Moisture- and cloud-related variables (cloud fraction, precipitation duration fraction, relative humidity, precipitation) form one cluster; vapour pressure, air temperature and vapour-pressure deficit form a second; solar flux and wind speed contribute additional, partly independent information; solid precipitation has negligible importance.

Figure 13 summarises the SHAP contributions of the edaphoclimatic group ($\phi_{edapho}$) by region. Each boxplot shows, for a given edaphoclimatic region, the distribution of $\phi_{edapho}$ across all municipality–year observations belonging to that region. Values to the right of the zero line indicate that, conditional on year and weather, the edaphoclimatic label tends to *increase* the predicted yield relative to the model baseline; values to the left indicate a tendency to *decrease* it. The width of each box reflects how heterogeneous this effect is within the region.

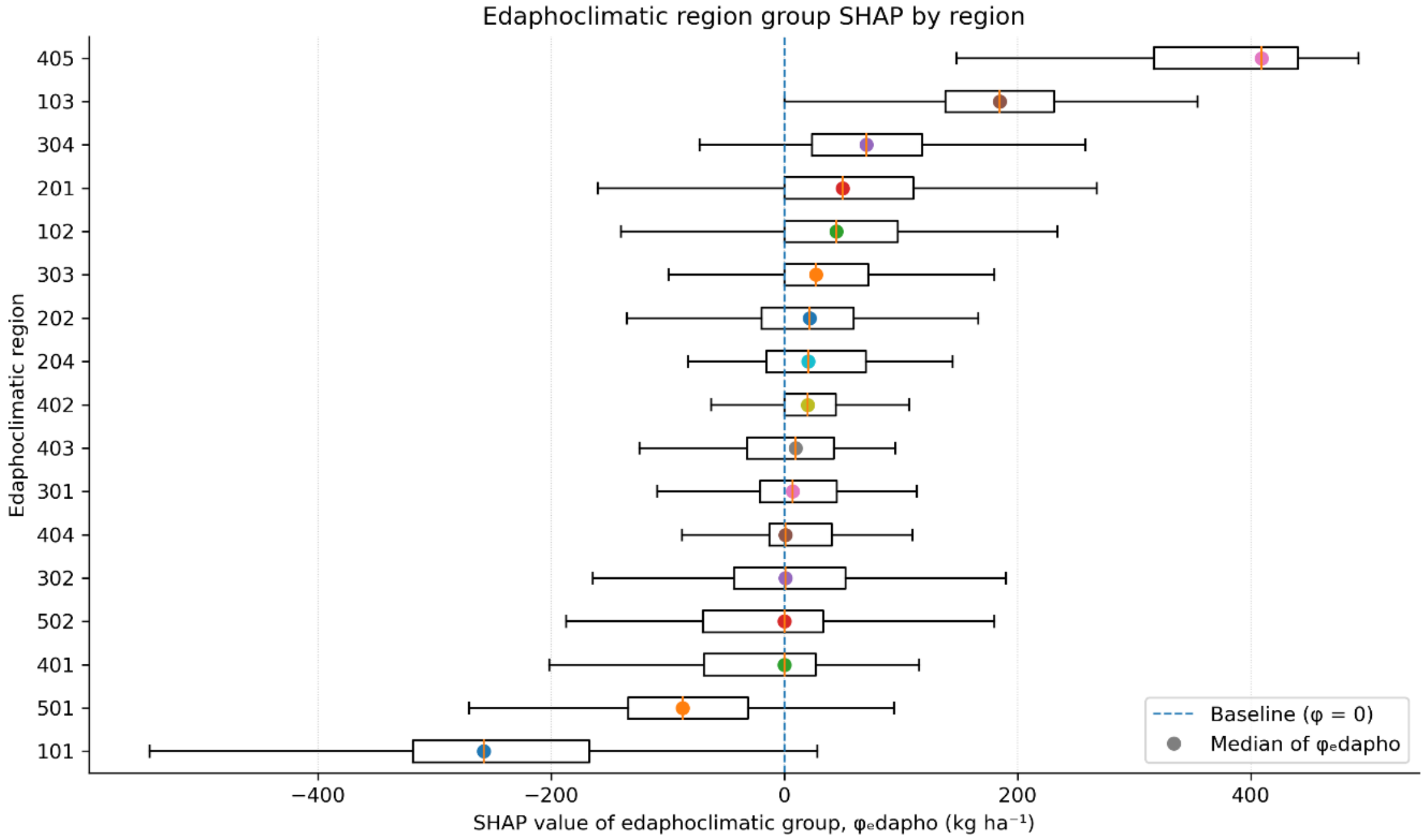

Figure 13: SHAP contributions of the edaphoclimatic group by region. Boxplots show the distribution of SHAP values for the edaphoclimatic feature group $(\phi_{\text{edapho}})$ across municipality–year observations in each edaphoclimatic region. The vertical dashed line marks zero contribution. Positive values indicate that, conditional on year and weather, the regional label increases the predicted yield relative to the model baseline, whereas negative values indicate a downward adjustment. The spread of each box reflects within-region heterogeneity; for example, Region 101 (largely corresponding to Rio Grande do Sul) shows wide variability in $\phi_{\text{edapho}}$, consistent with strong interannual and spatial contrasts in yield response.

Region 101, which largely corresponds to Rio Grande do Sul, exhibits the widest spread of $\phi_{edapho}$, indicating that the marginal contribution associated with this edaphoclimatic label varies strongly across municipality–year combinations. This pattern is consistent with the combination of large interannual yield variability in that state and non-linear interactions between the regional background and the climate and year predictors. Other regions display narrower boxes and more tightly centred medians, suggesting more stable regional adjustments once year and weather are accounted for.

More broadly, the fact that the edaphoclimatic group exerts a non-negligible and directionally coherent effect supports its role as a proxy for spatially structured factors not explicitly included in the dataset (e.g. soil properties, management practices, and cultivar choices). Importantly, the ordering of regions by median $\phi_{edapho}$ does not need to coincide with their ranking by observed mean yield, because SHAP values quantify the marginal contribution of the edaphoclimatic label after conditioning on the other predictors, rather than raw yield levels. In this sense, the edaphoclimatic label is capturing how the model adjusts yields up or down around the national baseline for each region, rather than reproducing simple differences in average productivity.

## 3.6 Strengths, limitations, and implications for transferability

Because the approach is input-frugal – using weather as the only time-varying signal plus simple static covariates – it is lightweight to deploy and broadly transferable across crops and regions. Here, "transferable" refers to the framework and its low-burden input requirements rather than to a single trained model. Applying the approach to a new crop or geography would typically require re-training (or fine-tuning) on local yield and meteorology, so that differences between climatic regimes, as well as soil–management contexts, are learned from data rather than assumed. This frugality does not come at the expense of accuracy: under an end-of-season, national-scale evaluation aligned to standard rRMSE reporting, our Transformer model attains performance that is on par with recent Brazil benchmarks, including data-rich pipelines that incorporate remote sensing and engineered agro-climatic indices (von Bloh *et al.*, 2023).

At the same time, its weather-centric design is forecast-ready: it can ingest operational (ensemble) weather forecasts and thereby capitalise on continuing improvements in weather-forecast skill to deliver earlier, more reliable updates for stakeholders. This contrasts with pipelines that depend on satellite imagery and/or field-based survey (interview-driven assessments), which typically entail greater latency, uneven coverage and higher operational cost and, in many cases, additional feature-engineering and preprocessing complexity (Jones *et al.*, 2003; Ma *et al.*, 2021; von Bloh *et al.*, 2023). In this sense, the framework targets a favourable accuracy–complexity trade-off: competitive national errors with materially simpler and more readily operationalised inputs.

The SHAP analysis of the reference Transformer model is consistent with this design choice: the crop-year covariate captures much of the long-run signal, while weather and the edaphoclimatic label provide systematic adjustments around that trend, indicating that the restricted input set is nevertheless sufficiently informative for the model to represent both technological progress and shorter-term climate-driven deviations.

Limitations follow directly from these design choices. With weather-only inputs, the model cannot capture non-weather shocks (for example pest or disease outbreaks) unless they manifest indirectly through the weather–yield pathway. Additionally, because the approach deliberately avoids crop-specific inputs and represents non-weather factors via an agro-environmental static label, it is not designed to anticipate structural changes as crop frontiers expand, or management practices and cultivars evolve. As a consequence, maintaining accuracy under shifting production regimes may require periodic retraining (or recalibration) as new seasons become available, and performance monitoring to detect regime changes that are not explained by weather alone.

A further source of avoidable mismatch is the use of a single national soybean planting calendar. Brazil exhibits marked regional heterogeneity in sowing dates and crop development, so aligning weather sequences to a fixed calendar can blur the correspondence between climatic conditions and phenological stages. Incorporating region-specific (and ideally year-specific) planting-date expectations would likely improve phase alignment and could yield measurable accuracy gains without compromising the framework's frugal philosophy. This could be implemented using publicly available regional sowing windows (rather than field-level data), preserving operational feasibility.

Relatedly, unlike remote-sensing-augmented approaches (e.g., those using vegetation indices), the model does not directly observe within-season canopy dynamics or management signals, which may explain performance gains reported in more data-intensive pipelines under some settings (Bloh, von *et al.*, 2023).

SHAP results also show that a substantial share of predictive power is associated with the historical yield trend encoded in the crop-year input, with weather and regional context refining, rather than overturning, this baseline. This diagnostic reinforces the importance of performance monitoring under regime shifts: if the technological trajectory or management baseline changes abruptly, the learned trend-dependent component of the model may become misaligned with the evolving system, and extrapolations beyond the observed period should be interpreted with appropriate caution (including consideration of updated edaphoclimatic mappings where relevant).

From a food-security perspective, the combination of:

- Frugal, reproducible inputs.
- Strong national-scale accuracy.
- Demonstrable in-season skill.

suggests that the framework can contribute to the broader aim of providing timely, spatially resolved yield information in support of Sustainable Development Goal 2, Target 2.C, on stable food-commodity markets and access to information. The contextual benchmarking reinforces this relevance: achieving competitive end-of-season national errors using operationally plausible inputs strengthens the case for deployment in information-constrained settings where satellite pipelines or survey-based systems may be delayed or inconsistent. The SHAP-based decomposition further supports this role by showing that the model's predictions reflect interpretable contributions from year, weather and edaphoclimatic region, which can be communicated to practitioners and policymakers and used to diagnose how the framework is functioning when transferred to new crops or geographies.

# 4 Conclusion

We set out to develop an architecture-agnostic deep learning framework for yield forecasting that is readily transferable across crops and regions/countries by restricting inputs to operationally plausible data streams. The resulting framework is intentionally frugal in inputs: weather is the only time-varying signal, paired with a simple agro-environmental label to encode stable regional context. Under a common fusion scheme and identical inputs, the Transformer model ranked at the top of the national-scale benchmark, and we therefore adopted it as the reference implementation for ablations, in-season variants and explainability.

Across LOYO years for Brazilian soybean (2001–2020), the reference Transformer achieved a national RMSE of 149 kg $ha^{-1}$ (rRMSE ≈ 5.3%) and a 47.6% error reduction relative to a five-year moving-average baseline. In-season accuracy improved monotonically as within-season information accumulated, reaching ~49–50% lower error by February, supporting timely use in procurement, logistics and risk management. At sub-national scale, error magnitudes were heterogeneous, but state-level trajectories showed no evidence of systematic temporal drift, while municipality-level diagnostics indicated predominantly positive and spatially coherent tracking of interannual variability, supporting the interpretation that national skill is not primarily an artefact of offsetting regional biases.

Key takeaways are as follows:

- Under identical inputs and a common fusion scheme, sequence encoders consistently outperform non-sequential baselines, with the Transformer and LSTM forming the top tier at national scale.
- Forecast skill strengthens in-season as weather information accrues, enabling earlier decision support relative to a farmer-style moving-average baseline.
- The agro-environmental context label and spatial instance expansion provide additive benefits while keeping the pipeline lightweight and transferable.
- Sub-national diagnostics show residual heterogeneity but no evidence of systematic temporal drift, reducing concern that national skill is driven primarily by regional error cancellation.

Explainability diagnostics further clarify how the reference model leverages the available signals. SHAP-based analyses indicate that the crop-year covariate captures most of the long-run yield trajectory, while within-season weather features and the edaphoclimatic label primarily refine interannual deviations around this trend. The dominant weather contributions arise from coherent groups of variables (notably moisture/cloud-related predictors and thermal-demand indicators), consistent with agronomic expectations and with the interpretation that correlated predictors act as overlapping signals rather than isolated causal drivers. More broadly, these diagnostics provide a transparent, operationally interpretable decomposition of trend versus weather-driven deviations, showing that the crop-year covariate captures the long-run trajectory while weather and edaphoclimatic context refine interannual departures around that baseline. This integrated treatment of trend simplifies pre- and post-processing relative to detrend/retrend workflows, while still making clear that, as in any trend-aware forecasting approach, model updating is needed when underlying production regimes shift.

Two design choices support portability: (i) the agro-environmental context label, which reduces municipality-level error by conditioning the mapping on production-environment context, and (ii) spatial instance expansion, included as a robustness device so the template remains stable when training data are sparse, an expected reality when extending to new crops or regions. Because the method relies on routine weather inputs, it can ingest operational weather forecasts and thereby support earlier, forecast-driven yield updates in-season (and potentially ahead of the season) without requiring additional data streams or heavy pipelines. Limitations include insensitivity to non-weather shocks, reliance on a simplified planting-calendar representation, and the need for periodic retraining or recalibration under evolving technological and management regimes. Future work will apply this template to other crops and geographies, and develop probabilistic, forecast-driven outputs to support operational decision-making at scale.

# 5 Acknowledgements and contributions

**Acknowledgment:**

This study was financed in part by the Coordenação de Aperfeiçoamento de Pessoal de Nível Superior—Brazil (CAPES)—Finance Code 001 - (Proc. 88881.846220/2023-01), Fundação de Amparo à Pesquisa do Estado da Bahia (FAPESB) (Proc. 493/2022), and the National Council for Scientific and Technological Development (CNPq, Brazil), through Finance Code 001. Erick G. Sperandio Nascimento is a CNPq technological development fellow (Proc. 308963/2022-9). Prashant Kumar acknowledges the support received through the UKRI (NERC, EPSRC, AHRC) funded RECLAIM Network Plus (EP/W034034/1). The authors also thank the Manufacturing and Technology Integrated Campus—SENAI CIMATEC, the Surrey Institute for People-Centred AI, and the Global Centre for Clean Air Research (GCARE) at the University of Surrey for their valuable support.

**Author Contributions:**

Conceptualization, F.D.d.C.M., P.K., and E.G.S.N.; methodology, P.K., and E.G.S.N.; data curation, F.D.d.C.M.; software, F.D.d.C.M.; formal analysis, F.D.d.C.M. and E.G.S.N.; original draft preparation, F.D.d.C.M.; writing—review and editing, F.D.d.C.M., P.K. and E.G.S.N.; supervision, E.G.S.N. All authors have read and agreed to the published version of the manuscript.